\documentclass[journal,twocolumn]{IEEEtran}
\usepackage{bm}
\usepackage[english]{babel}
\usepackage{amsmath,amsthm,cases,amsfonts,amssymb,mathrsfs}
\usepackage{epsfig, multirow,  cite, txfonts}
\usepackage{subfigure}

\usepackage{graphicx}
\usepackage{caption}
\usepackage{color,hyperref}
\newcommand{\argmin}{\mathop{{\rm arg}\min}}
\newcommand{\argmax}{\mathop{{\rm arg}\max}}
\newcommand{\mbf}[1]{\mathbf{#1}}
\newcommand{\mbb}[1]{\mathbb{#1}}

\newcommand{\mcal}[1]{\mathcal{#1}}
\DeclareMathOperator{\sign}{sign}
\usepackage{makecell}

\ifCLASSINFOpdf
\else
\fi

\definecolor{darkblue}{rgb}{0.0,0.0,1.0}
\hypersetup{colorlinks,breaklinks,
            linkcolor=darkblue,urlcolor=darkblue,
            anchorcolor=darkblue,citecolor=darkblue}
\newtheorem{definition}{Definition}

\begin{document}

\title{Modal Regression based Atomic Representation for Robust Face Recognition}

\author{Yulong~Wang,~\IEEEmembership{}
        Yuan Yan~Tang,~\IEEEmembership{Life Fellow, IEEE},
        Luoqing~Li,~\IEEEmembership{}
        and~Hong~Chen \IEEEmembership{}

\thanks{
Y. Wang  is with the School of Information Science and Engineering,
Chengdu University, Chengdu, 610106, China (e-mail: wangyulong6251@gmail.com).
Y. Y. Tang is with the Faculty of Science and Technology,
University of Macau, Macau 999078, China (e-mail: yytang@umac.mo).
L. Li is with the Faculty of Mathematics and Statistics,
Hubei University, Wuhan 430062, China (e-mail: lilq@hubu.edu.cn).
H. Chen is with the Department of Computer Science and Engineering,
University of Texas, Arlington, TX 76019, USA (e-mail: chenh@mail.hzau.edu.cn).}
}

\maketitle

\begin{abstract}
Representation based classification (RC) methods such as sparse RC (SRC)
have shown great potential in face recognition in recent years.
Most previous RC methods are based on
the conventional regression models, such as lasso regression,
ridge regression or group lasso regression.
These regression models essentially impose a predefined assumption on the distribution
of the noise variable in the query sample,
such as the Gaussian or Laplacian distribution.
However, the complicated noises in practice may violate the assumptions
and impede the performance of these RC methods.
In this paper, we propose a \emph{modal regression based atomic representation
and classification} (MRARC) framework to alleviate such limitation.
Unlike previous RC methods, the MRARC framework
does not require the noise variable to follow any specific predefined distributions.
This gives rise to the capability of MRARC in handling various complex
noises in reality.
Using MRARC as a general platform, we also develop four novel RC methods
for unimodal and multimodal face recognition, respectively.
In addition, we devise a general optimization algorithm for
the unified MRARC framework based on
the alternating direction method of multipliers (ADMM)
and half-quadratic theory.
The experiments on real-world data validate the efficacy
of MRARC for robust face recognition.
\end{abstract}


\IEEEpeerreviewmaketitle

\section{Introduction}
\label{sec:Introduction}

\IEEEPARstart{R}epresentation-based classification (RC) methods
have drawn intensive interest and shown great potential in face recognition
in recent years \cite{Wright09,Elhamifar12,ZhangLei11,HeR11}.
An appealing merit of RC methods is that they can exploit
the subspace structure of data in each class for classification.
Concretely, RC methods are based on the observation that
many real-world data in a class often approximately lie in a low-dimensional subspace,
such as face images of a subject under varying illumination \cite{Basri03}
and hand-written digit images with distinct rotations
and translations \cite{Hastie98}.

In the past decades, various RC methods have been proposed
for face recognition (FR).
Inspired by the success of lasso regression in compressed sensing,
Wright \emph{et al.} \cite{Wright09} first developed
the sparse RC (SRC) method for face recognition.
To improve the efficiency of SRC, Zhang \emph{et al.} \cite{ZhangLei11} put forward the  collaborative RC (CRC) approach by using the ridge regression to
compute the representation vector.
To exploit the block structure of the dictionary,
Elhamifar and Vidal \cite{Elhamifar11} proposed a block sparse RC (BSRC)
method by utilizing the group lasso regression for FR.
Zhang \emph{et al.} \cite{Zhang12} proposed a nonlinear extension of SRC by
incorporating the kernel trick into  SRC.
Shekhar \emph{et al.} \cite{Shekhar14} developed a joint sparse RC (JSRC) method
for multimodal face recognition by utilizing the correlation information
among distinct modalities.
More recent advances on RC methods can be found in the references \cite{YangM13a,HeR14,YangJian17,Me15TCYB,Me16ICPR}.
Previous RC methods are devised separately based on different motivations.
In our previous work \cite{Me15TCYB}, we developed a unified framework termed as atomic representation-based classification (ARC).
We show that many important RC methods can be reformulated as special cases of ARC.

\begin{figure}
  \centering
   \subfigure[]{
   \nonumber
   \includegraphics[width=1.2cm]{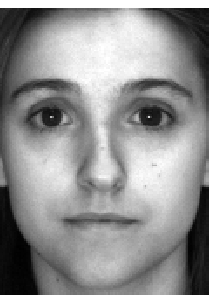}}
   \hspace{0.1in}
   \subfigure[]{
   \nonumber
   \includegraphics[width=1.2cm]{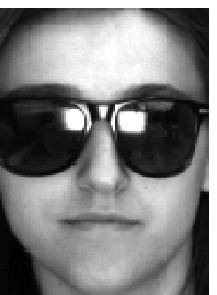}}
   \hspace{0.1in}
   \subfigure[]{
   \nonumber
   \includegraphics[width=1.2cm]{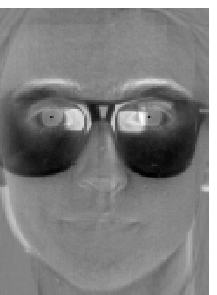}}
   \par
   \subfigure[]{
   \nonumber
   \includegraphics[width=6.0cm]{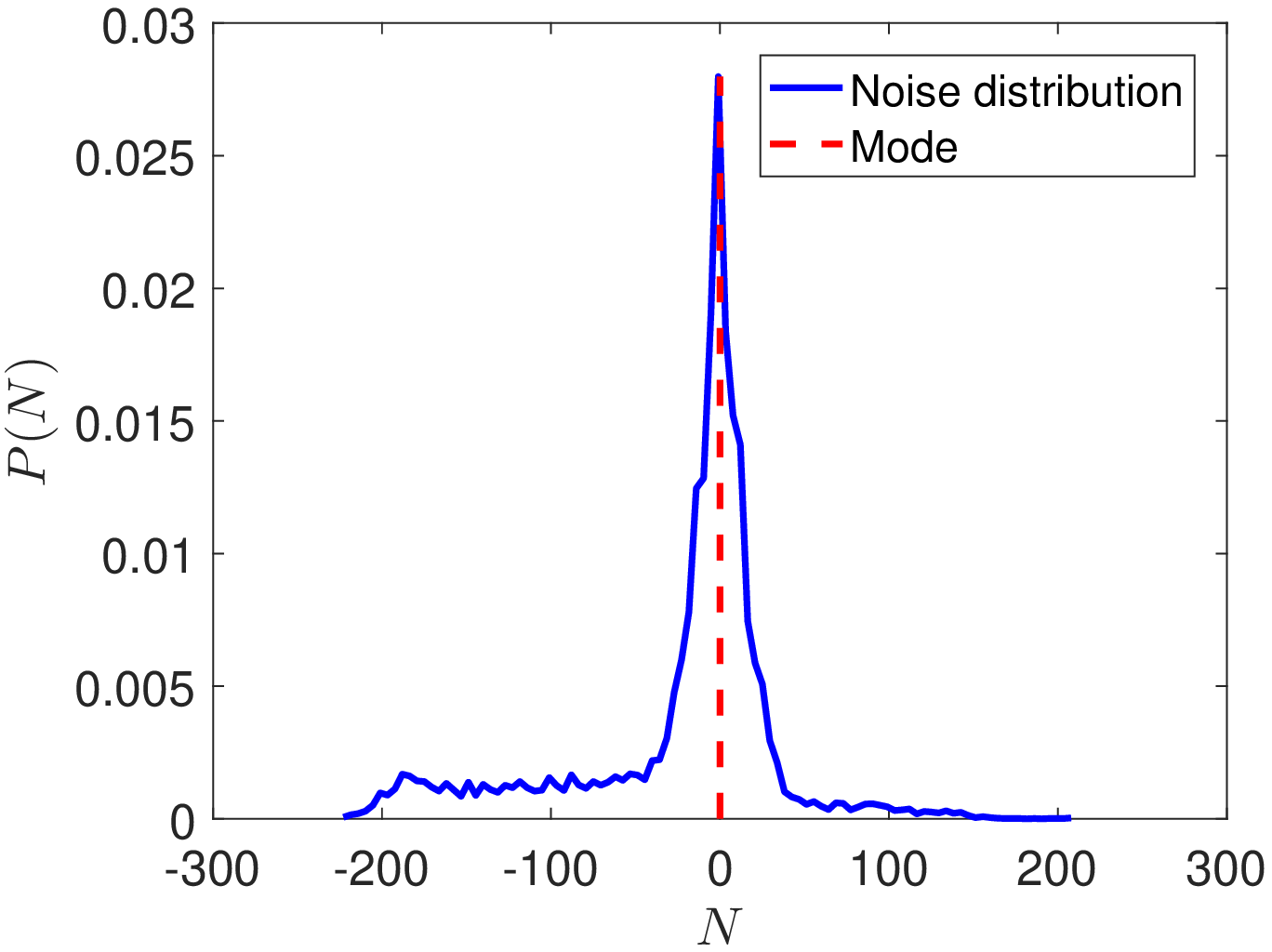}}
 \caption{\textbf{An example to show the mode of the noise variable.}
 (a) a noiseless facial image; (b) a facial image of the same subject with sunglasses; (c) the noise image, i.e., the difference between (b) and (a); (d) the noise distribution $p_{N}(t)$ of pixels in (c) and its mode, i.e.,
 $\text{mode}(N)=\argmax_{t\in\mbb{R}}p_{N}(t)$. }
\label{fig1}
\end{figure}

Despite the empirical success, most previous RC methods are based on the
conventional regression models, such as lasso regression \cite{Tibshirani96},
ridge regression \cite{Hoerl70} or group lasso regression \cite{Yuan06}.
These regression models in fact impose a predefined assumption on the distribution
of the noise variable, such as the Gaussian distribution or Laplacian distribution \cite{YangM13a,Erdogmus02a,Feng17}.
Such limitation may impede their performance
when the assumptions violate in the presence of complicated
noises in real-world face recognition.

In this paper, we propose to learn the representation vector based on
the modal regression and the atomic norm regularization.
The modal regression \cite{Sager82} aims to reveal the relationship between
the input variable and the response variable by regressing towards the
conditional mode function. For a continuous random variable $\xi$,
the mode is defined as the value at which its density function
$p_{\xi}(\cdot)$ attains its peak value, i.e.,
$\text{mode}(\xi)=\argmax_{t\in\mbb{R}}p_{\xi}(t)$.
For a set of observations, the mode is the value that appears most frequently.
Fig. \ref{fig1} shows the mode of the noise variable in a facial image
with sunglasses.
Previous research results \cite{Sager82,Zhou2016,Feng17}
have shown that one of the most appealing merits of modal
regression is its robustness to various complex noise,
including heavy-tailed noises, impulsive noises and outliers.
The novelties and contributions of this work are summarized as follows:
\begin{itemize}
\item[1.]
We develop a general unified framework termed as modal regression based atomic representation and classification (MRARC) for robust face recognition and reconstruction. Unlike previous RC methods, MRARC  does not require the noise variable to follow any specific predefined distributions.
This gives rise to its ability in handling the various complicated noises
in reality.

\item[2.]
Using MRARC as a general platform, we propose four novel
modal regression based RC methods by specifying distinct atomic sets for
unimodal and multimodal face recognition, respectively.

\item[3.]
We devise a general optimization algorithm for MRARC based on
the Alternating Direction Method of Multipliers (ADMM) \cite{Boyd12}
and half-quadratic theory \cite{Nikolova05}.
Thus, the algorithm can be applied to each method in MRARC.

\end{itemize}

The remainder of the paper is arranged as below.
Section \ref{sec:RelatedWork} briefly reviews previous related works.
In Section \ref{sec:MRARC}, we describe the proposed method.
Section \ref{sec:Experiments} presents the experiments on real-world databases.
Finally, Section \ref{sec:Conclusion} concludes.

\section{Related Work}
\label{sec:RelatedWork}

This section briefly introduces some representative RC methods.
Consider a classification problem with $K$ classes.
Let $\mbf{A}_k=\left[\mbf{a}_1^k, \mbf{a}_2^k, \cdots, \mbf{a}_{n_k}^k\right]\in\mathbb{R}^{m\times n_k}$
be a matrix of $n_k$ labeled training samples from the $k$-th class
for $k=1,2,\cdots,K$.
Define $\mbf{A}=\left[\mbf{A}_1,\mbf{A}_2,\cdots,\mbf{A}_K\right]$.
Table \ref{tab1} summarizes the key notations used in this paper.
Given the training data matrix $\mbf{A}$,
the goal is to correctly determine the label of any new test sample $\mbf{y}\in\mbb{R}^m$.

1) SRC (Sparse Representation based Classification) \cite{Wright09}:
The SRC method tries to seek the sparsest solution to the linear system of equations
$\mbf{y}=\mbf{A}\mbf{c}$ for classification. To this end, it first computes the
representation vector by solving the $\ell_1$ minimization problem
\begin{equation}\label{eq:SRC1}
\mbf{c}^{\text{SRC}}=\argmin_{\mbf{c} \in \mbb{R}^n} \|\mbf{c}\|_1
~~\text{s.t.}~~\mbf{y}=\mbf{A}\mbf{c},
\end{equation}
where the $\ell_1$ norm is defined as $\|\mbf{c}\|_1=\sum_{i=1}^n|c_i|$.
To deal with noise, SRC solves the following $\ell_1$ minimization problem
also known as lasso regression \cite{Tibshirani96}
\begin{equation}\label{eq:SRC2}
\mbf{c}^{\text{SRC}}=\argmin_{\mbf{c} \in \mbb{R}^n} \|\mbf{y}-\mbf{A}\mbf{c}\|_2^2+\lambda\|\mbf{c}\|_1,
\end{equation}
where $\lambda$ is a positive regularization parameter.

2) CRC (Collaborative Representation based Classification) \cite{ZhangLei11}:
To improve the efficiency of SRC, the CRC method computes the representation
vector by solving the $\ell_2$ norm based ridge regression problem
\begin{equation}\label{eq:CRC}
\mbf{c}^{\text{CRC}}=\argmin_{\mbf{c} \in \mbb{R}^n}
\|\mbf{y}-\mbf{A}\mbf{c}\|_2^2+\lambda\|\mbf{c}\|_2^2.
\end{equation}
The problem (\ref{eq:CRC}) has a closed-form solution, which
can be explicitly expressed as
$\mbf{c}^{\text{CRC}}=(\mbf{A}^T\mbf{A}+\lambda\mbf{I})^{-1}\mbf{A}^T\mbf{y}$.
Here $\mbf{I}$ denotes the identity matrix.

\begin{table}
\caption{Key notations used in this paper.}
\begin{center}
\normalsize
\begin{tabular}
{c|c}
\hline\cline{1-2}
\textbf{Notation}       &\textbf{Description}
\\ \hline
$K$                              &number of classes  \\
$m$                              &feature dimension  \\
$n$                                      &number of all training samples  \\
$\mbf{A}\in\mbb{R}^{m\times n}$
& matrix of all training samples\\ $\mbf{y}\in\mbb{R}^m$
& a new test sample\\
\hline\cline{1-2}
\end{tabular}
\end{center}
\label{tab1}
\end{table}

3) BSRC (Block Sparse Representation based Classification) \cite{Elhamifar11}:
This method suggests considering the block structure of training data.
It assumes that training samples in each class form a few blocks.
Let $\left\{\mcal{B}_l\right\}_{l=1}^L$ be a partition of $\mcal{I}=\{1,2,\cdots,n\}$,
i.e., $\bigcup_{l=1}^L\mcal{B}_l=\mcal{I}$ and $\bigcap_{l=1}^L\mcal{B}_l=\emptyset$.
It computes the representation vector based on group lasso regression
\begin{equation}\label{eq:BSRC}
\mbf{c}^{\text{BSRC}}=\argmin_{\mbf{c} \in \mbb{R}^n}
\|\mathbf{y}-\mathbf{A}\mathbf{c}\|_2^2+\lambda
\sum_{l=1}^L\left\|\mbf{c}_{\mcal{B}_l}\right\|_2,
\end{equation}
where $\mbf{c}_{\mcal{B}_l}$ denotes the subvector of $\mbf{c}$
with entries of $\mbf{c}$ indexed by  $\mcal{B}_l$.

After the representation vector $\mbf{c}$ is obtained,
RC methods compute the class-specific residuals for each class
$$
r_k(\mathbf{y})=\|\mathbf{y}-\mathbf{A}\delta_k(\mbf{c})\|_2,~k=1,2,\cdots,K,
$$
where $\delta_k(\mbf{c})\in\mbb{R}^n$ is the vector
that only keeps the nonzero entries of $\mbf{c}$
with respect to the $k$-th class \cite{Wright09}.
Finally, the test sample $\mathbf{y}$ is assigned to the class
yielding the minimal residual.

In our previous work \cite{Me15TCYB},
we have proposed a general unified framework called atomic
representation based classification (ARC).
Most RC methods can be reformulated as special cases of ARC by specifying
the atomic set. To review ARC, we first introduce the definition
of the atomic norm.
\begin{definition}\cite{Chandrasekaran12}\label{def:AN}
The atomic norm of $\mbf{x}$ with respect to an atomic set $\mcal{A}$
is defined by
$$
\|\mbf{x}\|_\mcal{A}:=\inf_{t>0}\left\{t: \mbf{x}\in t\cdot\text{conv}(\mcal{A})\right\},
$$
where $\text{conv}(\mcal{A})$ denotes the convex hull of the set $\mcal{A}$.
\end{definition}
Two typical examples of the atomic norm are the $\ell_1$ norm
and the nuclear norm \cite{Chandrasekaran12}.
The former induces sparsity for vectors while  the latter induces
low rankness for matrices.
The \emph{atomic representation} (AR) model is given by
\begin{equation}\label{eq:AR}
\mbf{c}^*=\argmin_{\mbf{c}\in \mbb{R}^n}\|\mbf{c}\|_{\mcal{A}}~~\text{s.t.}
~~\mbf{y}=\mbf{A}\mbf{c}.
\end{equation}
For noisy data $\mbf{y}$, we consider  the regularized AR model
\begin{equation}\label{eq:RAR}
\mbf{c}^*=\argmin_{\mbf{c}\in\mbb{R}^n}
\|\mbf{y}-\mbf{A}\mbf{c}\|_2^2+\lambda\|\mbf{c}\|_{\mcal{A}}.
\end{equation}
Then we compute the class-specific residual
$r_k(\mbf{y})=\|\mathbf{y}-\mathbf{A}\delta_k(\mbf{c}^*)\|_2$
for each class and assign $\mbf{y}$ to the class with the minimal residual.

\begin{figure}[!t]
  \centering
   \includegraphics[width=6.0cm]{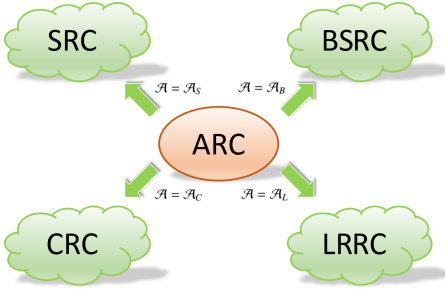}
 \caption{\textbf{As a general framework, ARC includes many RC methods as special cases.}}
\label{fig2}
\end{figure}

Most previous RC methods belong to ARC as special cases with
the specific atomic set $\mcal{A}$, as shown in Fig. \ref{fig2}.
For example, if we define
$\mcal{A}_S:=\left\{\pm\mbf{e}_i\right\}_{i=1}^n$
where $\mbf{e}_i\in\mbb{R}^n$ is a unit vector with its only nonzero
entry 1 in the $i$-th coordinate, we have
$\|\cdot\|_{\mcal{A}_S}=\|\cdot\|_1$ and ARC reduces to SRC \cite{Wright09}.
Similarly, BSRC \cite{Elhamifar12} also belongs to ARC.
Concretely, Let $\left\{\mcal{B}_l\right\}_{l=1}^L$ be a partition of $\mcal{I}=\{1,2,\cdots,n\}$ as mentioned before,
i.e., $\bigcup_{l=1}^L\mcal{B}_l=\mcal{I}$ and $\bigcap_{l=1}^L\mcal{B}_l=\emptyset$.
Denote $\mcal{B}_l^c=\mcal{I}-\mcal{B}_l$.
Define
$
\mcal{A}_B:=\bigcup_{l=1}^L
\left\{\mbf{a}\in\mbb{R}^n|~\left\|\mbf{a}_{\mcal{B}_l}\right\|_2=1,~
\left\|\mbf{a}_{\mcal{B}_l^c}\right\|_2=0\right\}.
$
It can be proved that
$\|\mbf{c}\|_{\mcal{A}_B}=\sum_{l=1}^L\left\|\mbf{c}_{\mcal{B}_l}\right\|_2$
and ARC reduces to BSRC by setting $\mcal{A}=\mcal{A}_B$.
It can be shown that CRC also belongs to ARC by using the atomic set
$\mcal{A}_C=:\left\{\mbf{a}\in \mbb{R}^n| ~\|\mbf{a}\|_2=1\right\}$.
Another example of ARC is the low-rank RC (LRRC) \cite{LiuG10} when
multiple  test samples are considered simultaneously.
Given $p$ test samples $\mbf{y}_1, \cdots, \mbf{y}_p\in\mbb{R}^m$,
we arrange them as columns of a matrix $\mbf{Y}=[\mbf{y}_1, \cdots, \mbf{y}_p]$.
The LRRC model looks for the representation matrix with the lowest rank by
\begin{equation}\label{eq:LRRC}
\min_{\mbf{C} \in \mbb{R}^{n\times p}}\|\mbf{C}\|_*
~~\text{s.t.}~~\mbf{Y}=\mbf{A}\mbf{C},
\end{equation}
$\|\mbf{C}\|_*$ denotes the nuclear norm of $\mbf{C}$,
i.e., the sum of singular values of $\mbf{C}$.
If we define the atomic set
$
\mcal{A}_L:=\left\{ \mbf{M} \in \mbb{R}^{n\times p}| ~
\text{rank}(\mbf{M})=1, \|\mbf{M}\|_F=1 \right\},
$
we have $\|\mbf{C}\|_{\mcal{A}}=\|\mbf{C}\|_*$
and ARC reduces to LRRC.


\section{Proposed Method}
\label{sec:MRARC}

In this section, we describe the proposed modal regression based
atomic representation and classification (MRARC) framework
for robust face recognition and reconstruction.

\subsection{Modal Regression}
\label{subsec:MR}

We first introduce some basic facts and analysis of modal regression
\cite{Sager82,Zhou2016,Feng17}.
Denote $X\in\mbb{R}^n$ and $Y\in\mbb{R}$ as the input and response random variables, respectively.
Consider the observation model
\begin{equation}\label{eq:observation}
Y=f^*(X)+N,
\end{equation}
where $f^*(X)$ denotes the unknown target function and
$N$ represents the noise term. The goal
of the regression problem is to approximate the unknown target function
$f^*(X)$.
Modal regression aims to recover the target function $f^*(X)$
by regressing towards the following modal regression function \cite{Sager82}.
\begin{definition}\label{def:MRFunction}
The modal regression function $f_{\textnormal{M}}:\mbb{R}^n\rightarrow\mbb{R}$
is defined as
$$
f_{\textnormal{M}}(\mbf{x}):=\text{mode}(Y|X=\mbf{x})
=\argmax_{t\in\mbb{R}}p_{Y|X}(t|X=\mbf{x}),~\mbf{x}\in\mbb{R}^n
$$
where $p_{Y|X}(t|X=\mbf{x})$ denotes the conditional density of $Y$ conditioned
on $X$.
\end{definition}
If we assume the mode of the conditional distribution of the noise
$N$ at any $x$ to be zero, i.e.,
$$
\text{mode}(N|X=\mbf{x}):=\argmax_{t\in\mbb{R}}p_{N|X}(t|X=\mbf{x})=0,
$$
there holds $f^*(X)=\text{mode}(Y|X)=f_{\text{M}}(X)$
according to Eq. (\ref{eq:observation}).
Here $p_{N|X}(t|X=\mbf{x})$ denotes the conditional density of
the noise $N$ conditioned on $X$.
Thus, under zero-mode noise assumption, we have $f^*(X)=f_{\text{M}}(X)$
and the target is converted to estimating the modal regression
function $f_{\text{M}}(X)$. To this end,
we introduce the modal regression risk \cite{Feng17} as below.
\begin{definition}\label{def:MRRisk}
For a measurable function $f:\mbb{R}^n\rightarrow\mbb{R}$, its modal regression
risk $R(f)$ is defined as
$$
R(f)=\int_{\mbb{R}^n} p_{Y|X}\left(f(\mbf{x})|X=\mbf{x}\right)d\rho_{X}(\mbf{x}),
$$
where $\rho_{X}(\mbf{x})$ denotes the marginal distribution of $X$.
\end{definition}
It can be proved that $f_{\text{M}}(\mbf{x})$ is the
minimizer of the risk $R(f)$ \cite{Feng17},
i.e., $f_{\text{M}}(\mbf{x})=\argmax_{f\in\mcal{M}}R(f)$,
where $\mcal{M}$ denotes the set of all measurable functions on $\mbb{R}^n$.
For any measurable function, denote $E_f=Y-f(X)$ as the error random variable.
According to Eq. (\ref{eq:observation}), there holds
$N=E_f+f(X)-f^*(X)$.
Then the density of $E_f$ can be formulated as
\begin{eqnarray*}
p_{E_f}(e)&=&\int_{\mbb{R}^n} p_{E_f|X}\left(e|X=\mbf{x}\right)d\rho_{X}(\mbf{x})\\
&=&\int_{\mbb{R}^n} p_{N|X}\left(e+f(\mbf{x})-f^*(\mbf{x})
|X=\mbf{x}\right)d\rho_{X}(\mbf{x}).
\end{eqnarray*}
Setting $e=0$, we have
\begin{eqnarray*}
p_{E_f}(0)&=&\int_{\mbb{R}^n} p_{N|X}\left(f(\mbf{x})-f^*(\mbf{x})|X=\mbf{x}\right)d\rho_{X}(\mbf{x})\\
&=&\int_{\mbb{R}^n} p_{Y|X}\left(f(\mbf{x})|X=\mbf{x}\right)d\rho_{X}(\mbf{x})\\
&=&R(f).
\end{eqnarray*}
Thus, the modal regression problem is converted to minimizing
the value of the density $p_{E_f}$ at $0$.
In reality, we often have only a finite samples $\{(\mbf{x}_i, y_i)\}_{i=1}^m$ and
the density $p_{E_f}$ is unknown.
For this reason, the Parzen window method \cite{Parzen62} is utilized to
estimate $p_{E_f}$ as below
\begin{equation}\label{eq:Parzen}
\hat{p}_{E_f}(e)=\frac{1}{m}\sum_{i=1}^m\mcal{K}(e-e_i),
\end{equation}
where $e_i=y_i-f(\mbf{x}_i)$ for $i=1,2,\cdots, m$ and $\mcal{K}(\cdot)$
denotes the general kernel function satisfying
$\mcal{K}(e)=\mcal{K}(-e)$.
Some common kernel functions include the Gaussian kernel
$\mcal{K}(e)=\frac{1}{\sqrt{2\pi}\sigma}\exp\left(-e^2/2\sigma^2\right)$
and the Epanechnikov kernel $\mcal{K}(e)=\frac{3}{4}\left(1-e^2\right)_+$,
where $(e)_+=e$ if $e\geq0$ and $(e)_+=0$ otherwise.
Then we have the estimator $\hat{f}_{\text{M}}$ of the modal regression function
$f_{\text{M}}$ as follows
\begin{equation}\label{eq:hatfm1}
\hat{f}_{\text{M}}=\argmax_{f\in\mcal{M}}\hat{p}_{E_f}(0)
=\frac{1}{m}\sum_{i=1}^m\mcal{K}(y_i-f(\mbf{x}_i)).
\end{equation}
Under the zero-mode noise assumption, there holds
$$
\hat{f}_{\text{M}}\approx f_{\text{M}}=f^*.
$$
Based on the analysis above, we can find that
modal regression does not require the noise to follow
any specific preset distributions such as the Gaussian distribution
required by some conventional regression models.
This makes modal regression attractive in handling various
complex noise in practice \cite{Feng17}.

\subsection{Modal Regression based Atomic Representation (MRAR)}
\label{subsec:MRARC}
Define the error vector $\mbf{e}=[e_1,e_2,\cdots,e_m]$ where
$e_i=y_i-f(\mbf{x}_i)$.
The problem (\ref{eq:hatfm1}) can be rewritten in the equivalent form
\begin{equation}\label{eq:hatfm2}
\hat{f}_{\text{M}}=\argmin_{f\in\mcal{M}}\mcal{L}_{\text{M}}(\mbf{e}),
\end{equation}
where $\mcal{L}_{\text{M}}(\mbf{e})$ denotes the
modal regression based loss function (MRLF)
\begin{equation}\label{eq:MRLF}
\mcal{L}_{\text{M}}(\mbf{e}):=m\left(1-\hat{p}_{E_f}(0)\right)=
\sum_{i=1}^m
\left(1-\mcal{K}(y_i-f(\mbf{x}_i))\right).
\end{equation}
For simplicity, consider the linear function $f(\mbf{x})=\mbf{x}^T\mbf{c}$
where $\mbf{c}\in\mbb{R}^n$ is the unknown coefficient vector.
Then $e_i=y_i-\mbf{x}_i^T\mbf{c}$ and $\mbf{e}=\mbf{y}-\mbf{X}\mbf{c}$,
where $\mbf{y}=[y_1,y_2,\cdots,y_m]^T$ and
$\mbf{X}=\left[\mbf{x}_1^T;\mbf{x}_2^T;\cdots;\mbf{x}_m^T\right]
\in\mbb{R}^{m\times n}$.
Incorporating the MRLF $\mcal{L}_{\text{M}}(\cdot)$ into
the regularized AR model (\ref{eq:RAR}),
we have the following \emph{modal regression based atomic representation}
(MRAR) model
\begin{equation}\label{eq:MRAR}
\min_{\mbf{c}\in \mathbb{R}^n}
\mcal{L}_{\text{M}}\left(\mbf{y}-\mbf{X}\mbf{c}\right)
+\lambda\|\mbf{c}\|_{\mcal{A}}.
\end{equation}
The problem above is difficult to tackle due to
the combination of the nonlinearity of the MRLF
and the abstract atomic norm regularization.
In addition, most previous optimization techniques
are originally proposed for RC methods with special atomic norms,
such as sparse representation.
They are difficult to be applied for the general MRAR framework.
In this paper, we devise an effective optimization algorithm
to implement the general MRAR model based on ADMM \cite{Boyd12}
and the half-quadratic (HQ) theory \cite{Nikolova05}.
We first introduce an auxiliary vector $\mbf{z}$ and
reformulate the objective function in Eq. (\ref{eq:MRAR}) as
\begin{equation}\label{eq:GeneralAR1}
\min_{\mbf{c}, \mbf{z}\in \mathbb{R}^n}\mcal{L}_{\text{M}}\left(\mbf{y}-\mbf{X}\mbf{z}\right)+
\lambda\|\mbf{c}\|_{\mcal{A}},
~~\text{s.t.}~~\mbf{c}=\mbf{z}.
\end{equation}
The augmented Lagrangian function of Eq. (\ref{eq:GeneralAR1}) is
\begin{equation}\label{eq:ARLagrange}
\mbf{L}(\mbf{c},\mbf{z},\bm{\Lambda})
=\mcal{L}_{\text{M}}\left(\mbf{y}-\mbf{X}\mbf{z}\right)+\lambda\|\mbf{c}\|_{\mcal{A}}
 +\frac{\mu}{2}\left\|\mbf{c}-\mbf{z}+\bm{\Lambda}/\mu\right\|_2^2+h(\bm{\Lambda}),
\end{equation}
where $h(\bm{\Lambda})=-\frac{1}{2\mu}\|\bm{\Lambda}\|_2^2$ is independent
of $\mbf{c}$ and $\mbf{z}$. Here $\bm{\Lambda}$ is the Lagrangian multiplier and $\mu>0$ denotes a penalty parameter.
Given the initialization of $\mbf{c}$, $\mbf{z}$ and $\bm{\Lambda}$,
we alternatively update each variable while fixing others in each iteration.

In the first step, we update $\mbf{c}$ while fixing $\mbf{z}$ and $\bm{\Lambda}$ by
\begin{equation}\label{eq:ARUpdatex}
\min_{\mbf{c}}\mbf{L}(\mbf{c},\mbf{z},\bm{\Lambda})
=\min_{\mbf{c}}\frac{1}{2}\|\mbf{c}-(\mbf{z}-\bm{\Lambda}/\mu)\|_2^2
+\frac{\lambda}{\mu}\|\mbf{c}\|_{\mcal{A}}.
\end{equation}

\noindent\rule{0.5\textwidth}{1.5pt}

\noindent \textbf{Algorithm 1}  Implementation of MRAR (\ref{eq:MRAR})

\vspace{-0.1in}
\noindent\rule{0.5\textwidth}{1pt}

\noindent {\bfseries Input:} $\mcal{A}$,  $\mbf{X}$, $\mbf{y}$, and $\lambda$.

\noindent {\bfseries Initialization:}
$i=0$, $\mbf{c}^{(0)}=\mbf{0}$, $\mbf{z}^{(0)}=\mbf{0}$, $\bm{\Lambda}^{(0)}=\mbf{0}$,
$\mu=10^{-1}$, $\varepsilon=10^{-7}$, $\text{maxIter}=10^5$.

\noindent \textbf{while} not converged and $i<\text{maxIter}$ \textbf{do}
\begin{itemize}
\item[1:] Update $\mbf{c}$ by the proximity operator
\begin{equation*}
\mbf{c}^{(i+1)}
=\Pi_{\mcal{A}}\left(\mbf{z}^{(i)}-\bm{\Lambda}^{(i)}/\mu;\lambda/\mu\right).
\end{equation*}

\item[2:] Update $\mbf{z}$  as $\mbf{z}^{(i+1)}$    by Eq. (\ref{eq:ARUpdatez}).

\item[3:] Update the Lagrange multiplier vector by
\begin{equation*}
\bm{\Lambda}^{(i+1)}
=\bm{\Lambda}^{(i)}+\mu\left(\mbf{c}^{(i+1)}-\mbf{z}^{(i+1)}\right).
\end{equation*}

\item[4:] Check the convergence conditions:
\begin{equation*}
  \left\|\mbf{c}^{(i+1)}-\mbf{z}^{(i+1)}\right\|_{\infty}<\varepsilon
  ~\text{and}~
  \left\|\mbf{c}^{(i+1)}-\mbf{c}^{(i)}\right\|_{\infty}<\varepsilon
\end{equation*}

\item[5:] $i\leftarrow i+1$
\end{itemize}
\noindent \textbf{end while}

\noindent{\bfseries Output:} $\mbf{c}^*=\mbf{c}^{(i)}$.

\noindent\rule{0.5\textwidth}{1pt}
The optimal solution of Eq. (\ref{eq:ARUpdatex}) can be written as
\begin{equation}\label{eq:ARUpdatex1}
\hat{\mbf{c}}=\Pi_{\mcal{A}}\left(\mbf{z}-\bm{\Lambda}/\mu;\lambda/\mu\right),
\end{equation}
where
$
\Pi_{\mcal{A}}(\mbf{z},\gamma)=\argmin_{\mbf{c}}
\frac{1}{2}\|\mbf{c}-\mbf{z}\|_2^2
+\gamma\|\mbf{c}\|_{\mcal{A}}
$
denotes the proximity operator with respect to  $\mcal{A}$ \cite{Chandrasekaran12}.
Here we introduce the proximity operators for some common atomic sets
\begin{itemize}
  \item $\Pi_{\mcal{A}_S}(\mbf{z},\gamma)=\sign(\mbf{z})\otimes\left(|\mbf{z}|-\gamma\mbf{1}\right)_+$,

  \item $\Pi_{\mcal{A}_C}(\mbf{z},\gamma)=
        \frac{(\|\mbf{z}\|_2-\gamma)_+}{\|\mbf{z}\|_2}\mbf{z}$,

  \item $\Pi_{\mcal{A}_B}(\mbf{z},\gamma)_{\mcal{B}_l}=
         \Pi_{\mcal{A}_C}(\mbf{z}_{\mcal{B}_l},\gamma), ~l=1,2,\cdots,L.
        $
\end{itemize}
Here $\sign(\mbf{z})$ denotes a vector of the sign of entries of $\mbf{z}$
and $\otimes$ represents the Hadamard product.
$(x)_+=x$ if $x\geq0$ and $(x)_+=0$ otherwise.
$\{\mcal{B}_l\}_{l=1}^L$ denotes the $L$ index sets
in BSRC aforementioned.
For the vector $\mbf{z}$, $\mbf{z}_{\mcal{B}_l}$ denotes the subvector of $\mbf{z}$ containing the entries indexed by the set $\mcal{B}_l$.

In the second step, we update the auxiliary variable $\mbf{z}$ while
fixing $\mbf{c}$ and $\bm{\Lambda}$ by
\begin{equation}\label{eq:ARUpdatez}
\hat{\mbf{z}}=\argmin_{\mbf{z}}\mcal{L}_{\text{M}}\left(\mbf{y}-\mbf{X}\mbf{z}\right)
+\frac{\mu}{2}\left\|\mbf{z}-(\mbf{c}+\bm{\Lambda}/\mu)\right\|_2^2.
\end{equation}
We optimize the problem in (\ref{eq:ARUpdatez})
based on the half-quadratic (HQ) theory \cite{Nikolova05}.
For the function $\phi(u)=1-\mcal{K}(u)$, there exists
a dual convex function $\psi(v)$ \cite{Nikolova05} such that
$$
\phi(u)=\inf\left\{\frac{1}{2}vu^2 +\psi(v),~v\in \mathbb{R}\right\},
$$
where the infimum is reached at
For the Gaussian kernel $\mcal{K}(u)=\frac{1}{\sqrt{2\pi}\sigma}
\exp\left(-u^2/2\sigma^2\right)$,
$\tau(u)=\frac{1}{\sigma^2}\mcal{K}(u)>0$.
Then the MRLF in Eq. (\ref{eq:MRLF}) is rewritten   as
\begin{eqnarray*}\label{eq:hq2}
\mcal{L}_{\text{M}}(\mbf{e})
&=&\min_{\mbf{w}\in\mbb{R}_+^m}\sum_{i=1}^m \frac{1}{2}w_ie_i^2+\sum_{i=1}^m\psi(w_i)\\
&=&\min_{\mbf{w}\in\mbb{R}_+^m}\frac{1}{2}
\left\|\mbf{w}^{\frac{1}{2}}\otimes \mbf{e}\right\|_2^2+\sum_{i=1}^m\psi(w_i),
\end{eqnarray*}
where $\mbf{w}=[w_1, w_2, \cdots, w_m]\in\mbb{R}_+^m$.

\newpage
\noindent\rule{0.5\textwidth}{1.5pt}

\noindent \textbf{Algorithm 2}  MRAR based Classification

\vspace{-0.1in}
\noindent\rule{0.5\textwidth}{1pt}

\noindent {\bfseries Input:} An atomic set $\mcal{A}$,
                             training samples
                             $\mbf{A}=[\mbf{a}_1,\mbf{a}_2,\cdots,\mbf{a}_n]$,
                             a test sample $\mbf{y}$,
                             and the parameter $\lambda$.

\noindent{\bfseries Output:} identity $(\mbf{y})$.

\begin{itemize}

\item[1:] Normalize the columns of $\mbf{A}$ to have unit Euclidean norm.

\item[2:] Learn the representation vector $\mbf{c}^*$ via MRAR (\ref{eq:MRAR}).

\item[3:] Calculate the residuals
$$
r_k(\mbf{y})=\mcal{L}_{\text{M}}\left(\mbf{y}-\mbf{A}
\delta_{k}(\mbf{c}^*)\right),~k=1,2,\cdots,K.
$$

\item[4:] Predict $\text{identity}(\mbf{y})=\argmin_{k} r_k(\mbf{y})$.

\end{itemize}

\noindent\rule{0.5\textwidth}{1pt}
Thus, the problem in (\ref{eq:ARUpdatez}) can be reformulated as
\begin{equation}\label{eq:ARUpdatezIT2}
\min_{\mbf{z},\mbf{w}}
\left\|\mbf{w}^{\frac{1}{2}}\otimes\left(\mbf{y}-\mbf{X}\mbf{z}\right)\right\|_2^2
+2\sum_{i=1}^m\psi(w_i)
+\mu\left\|\mbf{z}-(\mbf{c}+\bm{\Lambda}/\mu)\right\|_2^2.
\end{equation}
In light of the HQ theory \cite{Nikolova05}, the problem (\ref{eq:ARUpdatezIT2}) can be
tackled by the following alternate procedure
\begin{equation}\label{eq:HQupdatew}
  w_i=\tau\left(y_i-\mbf{x}_i^T\mbf{z}\right),~i=1,2, \cdots,m
\end{equation}
\begin{equation}\label{eq:HQupdatez}
\mbf{z} = \argmin_{\mbf{z}\in\mbb{R}^n}
\left\|\mbf{w}^{\frac{1}{2}}\otimes\left(\mbf{y}-\mbf{X}\mbf{z}\right)\right\|_2^2
+\mu\left\|\mbf{z}-(\mbf{c}+\bm{\Lambda}/\mu)\right\|_2^2.
\end{equation}
If the Gaussian kernel is used,
the scale parameter $\sigma$ is often determined empirically
$
\sigma =\left(\frac{1}{2m}\left\|\mbf{y}-\mbf{X}
\mbf{z}\right\|_2^2\right)^{\frac{1}{2}}
$\cite{Principe10}.
The problem (\ref{eq:HQupdatez}) has a closed-form solution
$$
\mbf{z}=\left(\mbf{X}^T\text{diag}(\mbf{w})\mbf{X}+\mu\mbf{I} \right)^{-1}
\left(\mbf{X}^T\text{diag}(\mbf{w})\mbf{y}+\mu\mbf{c}+\bm{\Lambda}\right),
$$
where $\text{diag}(\mbf{w})$ denotes a square diagonal matrix with the
elements of $\mbf{w}$ on the main diagonal.
The iterations above are guaranteed to converge according to
the HQ theory \cite{Nikolova05}.
Finally, the Lagrange multiplier vector $\bm{\Lambda}$ is updated by
$\bm{\Lambda}=\bm{\Lambda}+\mu\left(\mbf{c}-\mbf{z}\right)$.
Algorithm 1 summarizes the algorithm of MRAR.

\subsection{MRAR based Classification}
\label{subsec:MRARC}

In this section, we develop the general MRAR based classification (MRARC) framework
and some novel methods as special cases of MRARC.

Given the training samples $\mbf{A}=\left[\mbf{a}_1,\mbf{a}_2,\cdots,\mbf{a}_n\right]$
and a new test sample $\mbf{y}$,
the first step is to compute the optimal coefficient vector $\mbf{c}^*$
using the MRAR model
\begin{equation}\label{eq:MRAR3}
\min_{\mbf{c}\in \mathbb{R}^n}
\mcal{L}_{\text{M}}\left(\mbf{y}-\mbf{A}\mbf{c}\right)
+\lambda\|\mbf{c}\|_{\mcal{A}}.
\end{equation}
Secondly, we calculate the class-specific residuals for each class.
Unlike most previous methods using the $\ell_2$ norm,
we utilize the MRLF to calculate the residuals
$$
r_k(\mbf{y})=\mcal{L}_{\text{M}}\left(\mbf{y}-\mbf{A}\delta_{k}(\mbf{c}^*)\right),
~k=1,2,\cdots,K,
$$
where $\delta_k(\mbf{c}^*)$ denotes the vector only keeping the nonzero entries of $\mbf{c}^*$ with respect to the $k$-th class.
Finally, the test sample $\mbf{y}$ is assigned to
the class yielding the minimal residual.
Algorithm 2 summarizes the classification procedure of MRARC.

It is worth pointing out that MRARC is a general framework for
pattern classification. We can use it to devise new classification
methods by specifying the atomic set $\mcal{A}$.
Concretely, we refer to the MRARC with the atomic sets
$\mcal{A}_S$,  $\mcal{A}_B$, and $\mcal{A}_C$ as
MRSRC, MRBSRC and MRCRC for short, respectively.

\subsection{MRARC for Multimodal Data}

Assume that we have $M$ modalities and the corresponding
dimensions are $m_1,\cdots,m_M$.
Let $\mbf{y}^j\in\mbb{R}^{m_j}$ and $\mbf{A}^j\in\mbb{R}^{m_j\times n}$
be the test sample and dictionary in the $j$-th modality where $j=1,2,\cdots,M$.
For multimodal data, the MRAR model can be formulated as
\begin{equation}\label{eq:ITMAR}
\min_{\mbf{X} \in \mbb{R}^{n\times M}}
\sum_{j=1}^M\mcal{L}_{\text{M}}\left(\mbf{y}^j-\mbf{A}^j\mbf{C}(:,j)\right)
+\lambda\|\mbf{C}\|_{\mcal{A}},
\end{equation}
where $\mbf{C}(:,j)$ denotes the $j$-th column of the matrix  $\mbf{C}$ and
$\mcal{A}\subset\mbb{R}^{n\times M}$ is an atomic set of matrices.
To take advantage of the correlation information among multiple modalities,
we can use the joint sparsity inducing atomic set
$$
\mcal{A}_J:=\bigcup_{i=1}^n
\left\{ \mbf{M} \in \mbb{R}^{n\times M}| ~
\|\mbf{M}(i,:)\|_2=1, \mbf{M}(i',:)=\mbf{0}, i'\neq i \right\},
$$
where $\mbf{M}(i,:)$ denotes the $i$-th row of the matrix $\mbf{M}$.
We refer to the MRARC method using $\mcal{A}_J$ in Eq. (\ref{eq:ITMAR}) as
modal regression based joint sparse representation classification (MRJSRC).
The MRJSRC method encourages the representation vectors of a test data
 (i.e., columns of $\mbf{C}$)
in distinct modalities to have the same sparsity pattern and locations of nonzero entries.
The optimization problem in Eq. (\ref{eq:ITMAR}) can be tackled using the
similar way as Algorithm 2.
The main difference is that the proximity operator for the atomic set
$\mcal{A}_J$ is formulated as
$$
\Pi_{\mcal{A}_J}(\mbf{Z},\gamma)(i,:)
  =\frac{\left(\|\mbf{Z}(i,:)\|_2-\gamma\right)_+}{\|\mbf{Z}(i,:)\|_2}
  \mbf{Z}(i,:),
$$
for $i=1,2,\cdots,n$.
Once the solution $\mbf{C}^*$ of Eq. (\ref{eq:ITMAR}) is obtained,
the class-dependent residuals are computed by
$$
r_k(\mbf{y})=\sum_{j=1}^M\mcal{L}_{\text{M}}
\left(\mbf{y}^j-\mbf{A}^j\delta_k\left(\mbf{C}^*(:,j)\right)\right),
~k=1,2,\cdots,K.
$$
Finally, the multimodal test sample $\left\{\mbf{y}^j\right\}_{j=1}^M$ is assigned to
the class yielding the minimal residual.


\section{Experiments}
\label{sec:Experiments}
In this section, we evaluate the efficacy of the proposed MRARC
framework for unimodal and multimodal face recognition against various
noises.

\begin{figure*}[!t]
  \centering
   \subfigure[]{
    \nonumber
   \includegraphics[height=1.8cm]{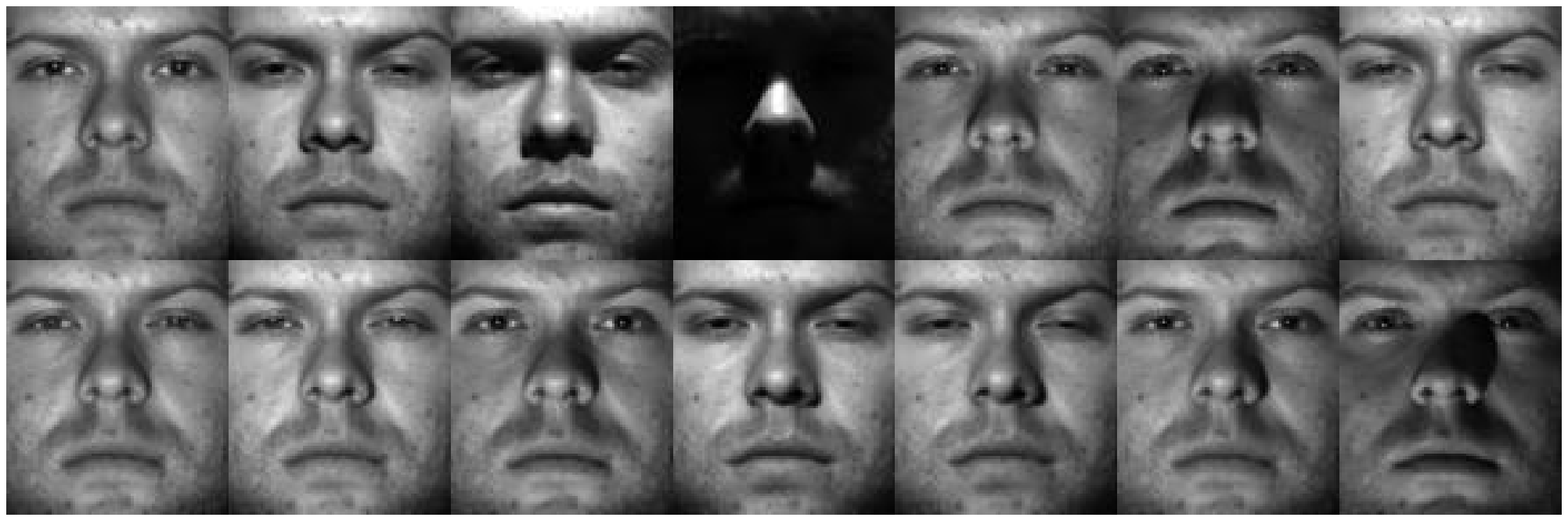}}
   \subfigure[]{
    \nonumber
   \includegraphics[height=1.8cm]{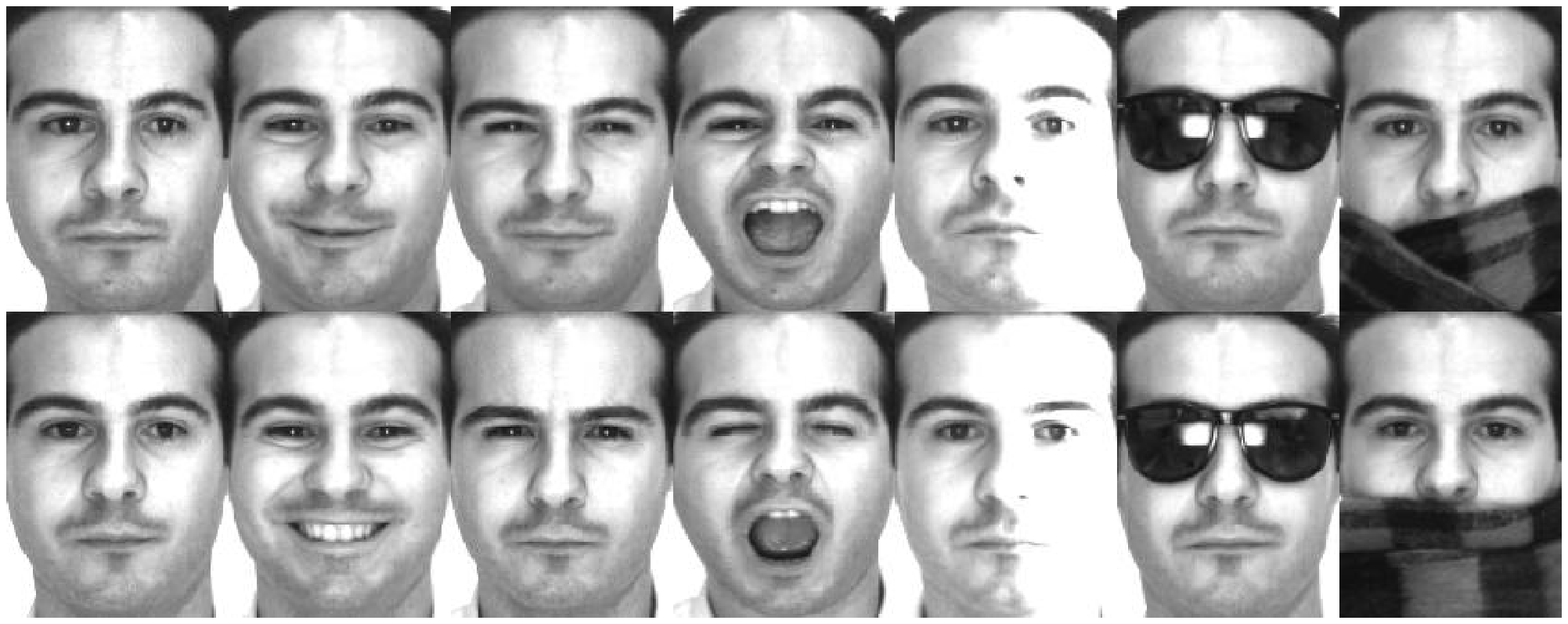}}
   \subfigure[]{
   \nonumber
   \includegraphics[height=1.8cm]{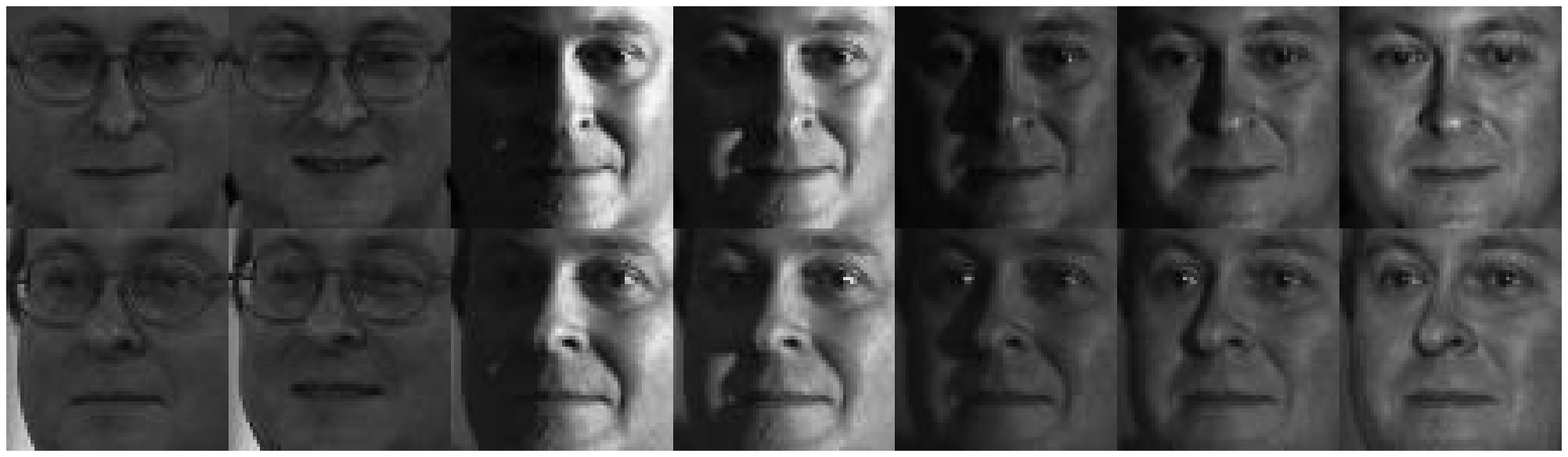}}
 \caption{\textbf{Sample images from different databases.}
 (a) the Extended Yale B database; (b) the AR database.
 (c) the CMU PIE database.
 }
\label{fig3}
\end{figure*}

Experiments are conducted on four public available databases, i.e.,
the Extended Yale B database (EYaleB) \cite{Lee05}, the AR database \cite{Martinez98},
the CMU MoBo database \cite{Gross01} and the the CMU PIE database \cite{Sim03}.
Fig. \ref{fig3} shows some sample images in these databases and
Table \ref{tab2} depicts the details of them,
including the number of classes, data dimension and the number of instances.
For unimodal face recognition, we compare MRSRC, MRBSRC
and MRCRC in the MRARC framework with the three typical
methods SRC, BSRC and CRC in the ARC framework.
The linear regression-based classification (LRC) \cite{Naseem10} approach
is used as the baseline. For multimodal face recognition,
we compare the proposed MRJSRC in MRARC with the JSRC \cite{Shekhar14} method.
For ARC and MRARC methods, the regularization parameter $\lambda$ is tuned by
searching a discrete set $\left\{10^{-5}, 10^{-4}, \cdots, 1,10\right\}$
to achieve their best performance as possible.
For MRARC methods, we set the penalty parameter $\mu=0.1$ in all experiments.


\begin{table}
\caption{Description of the used public databases.}
\begin{center}
\normalsize
\begin{tabular}
{c|c|c|c}
\hline\cline{1-4}
\textbf{Database}       &\#\textbf{Class} &\#\textbf{Dimension} &\#\textbf{Instance}
\\ \hline
Extended Yale B         &38               &$192\times 168$      &2,432  (images) \\
AR                      &100              &$165\times 120$      &2,600  (images) \\
CMU MoBo                &24               &$40\times 40$        &96    (videos) \\
CMU PIE                 &68               &$64\times 64$        &41,368    (images) \\
\hline\cline{1-4}
\end{tabular}
\end{center}
\label{tab2}
\end{table}

\subsection{Face Recognition With Occlusion}
\label{subsec:RecogOcc}

In this subsection, we conduct five different experiments to
analyze the performance of proposed methods for face recognition
and reconstruction against occlusions.

\begin{figure}[!t]
  \centering
   \includegraphics[width=8.0cm]{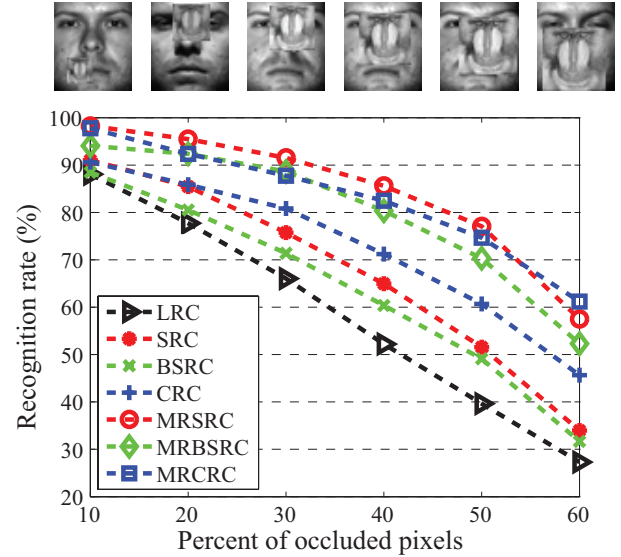}
 \caption{\textbf{Recognition rates versus percent of occluded pixels}.
 Recognition rates as a function of the percent of occluded pixels
 using the Extended Yale B Database.}
\label{fig4}
\end{figure}

\textbf{Experiment 1--Effect of percent of occlusion:}
In the first experiment, we evaluate the performance of
the competing methods against different levels of random occlusions.
For each test image, a random square region is occluded by a baboon image,
as shown in Fig. \ref{fig4}.
The Extended Yale B database is used for the experiment and the images
are resized to $48\times42$ for efficiency.
In the literature of RC methods \cite{Wright09}, many researchers manually chose
a subset of images in the database with normal or moderate light conditions for
training and only test images have extreme light conditions.
However, in real-world scenarios both training and testing images
may have different light conditions, including moderate and extreme light conditions.
For this reason,
we randomly select half of images (32 images) per subject
for training and the rest for testing.
Fig. \ref{fig4} shows the recognition rates of various methods
as a function of the percent of occlusion.
From Fig. \ref{fig4}, it can be seen that most competing methods
achieve high recognition rates when the occlusion level is low.
However, as the occlusion level increases, the recognition rates
of LRC, and the three methods (SRC, BSRC and CRC) in ARC drop rapidly.
In contrast, the three methods (MRSRC, MRBSRC and MRCRC) in MRARC outperform
other  methods in different occlusion levels.

\begin{figure*}[!t]
  \centering
   \subfigure[]{
    \nonumber
   \includegraphics[width=5.0cm]{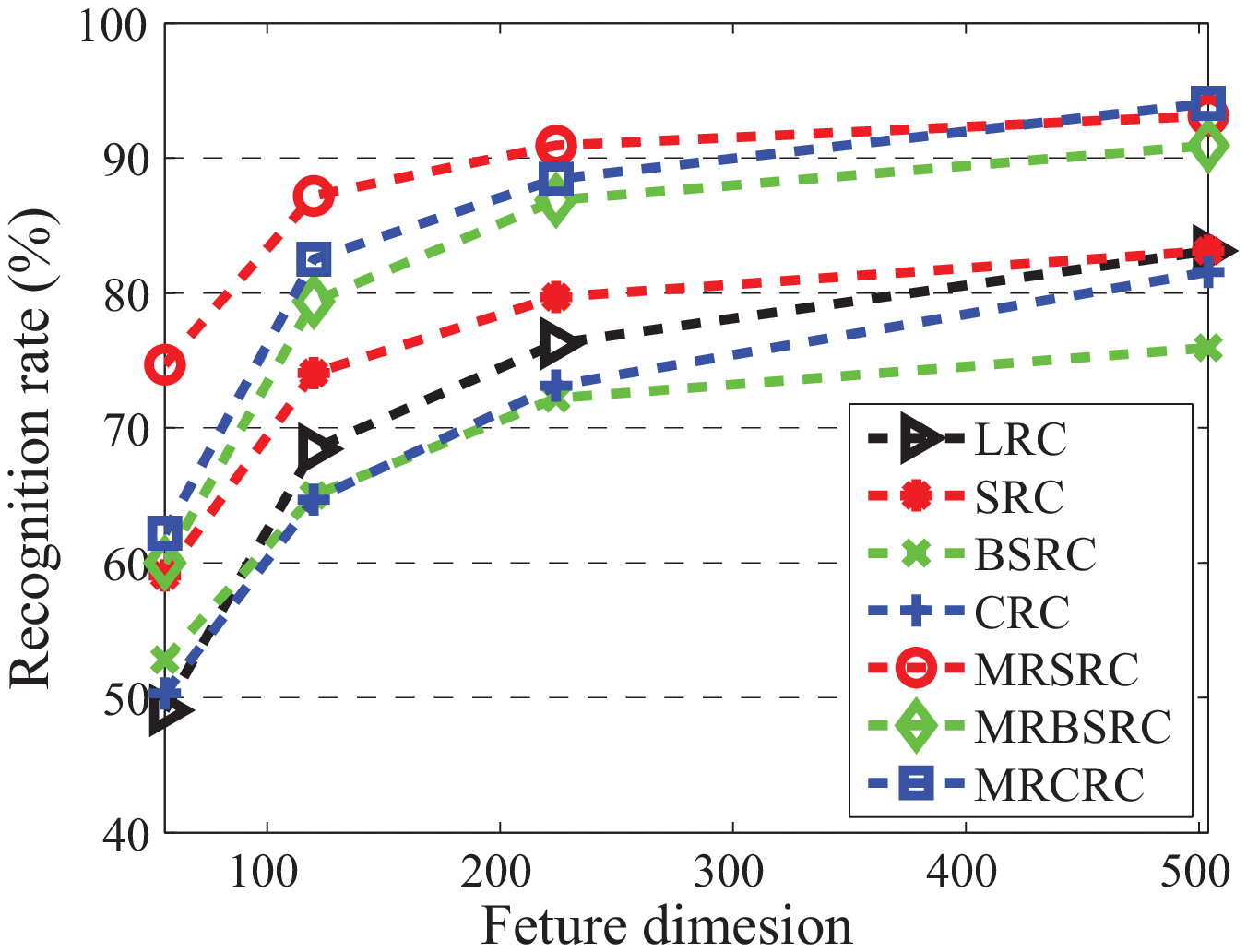}}
   \hspace{0.1in}
   \subfigure[]{
    \nonumber
   \includegraphics[width=5.0cm]{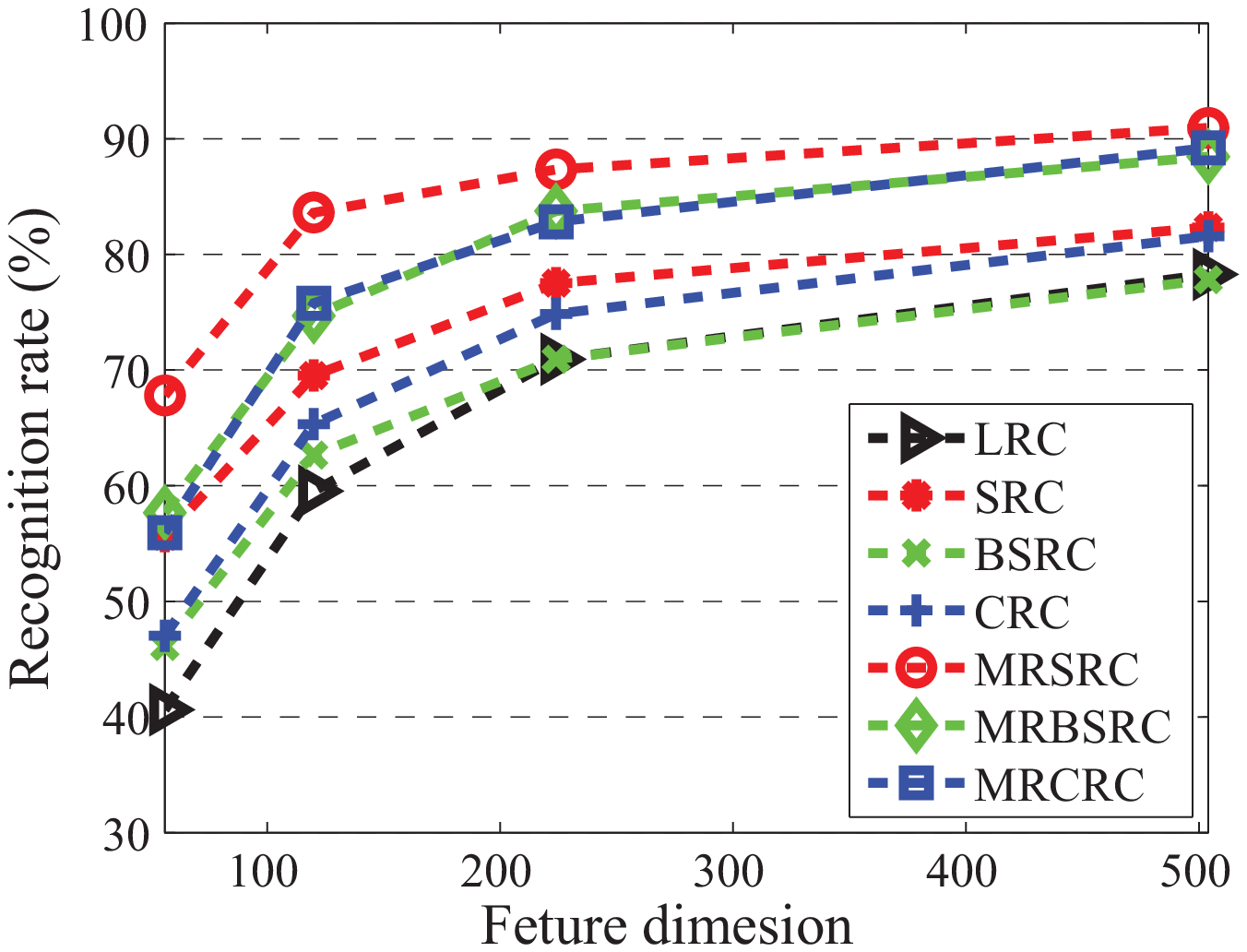}}
   \hspace{0.1in}
   \subfigure[]{
    \nonumber
   \includegraphics[width=5.0cm]{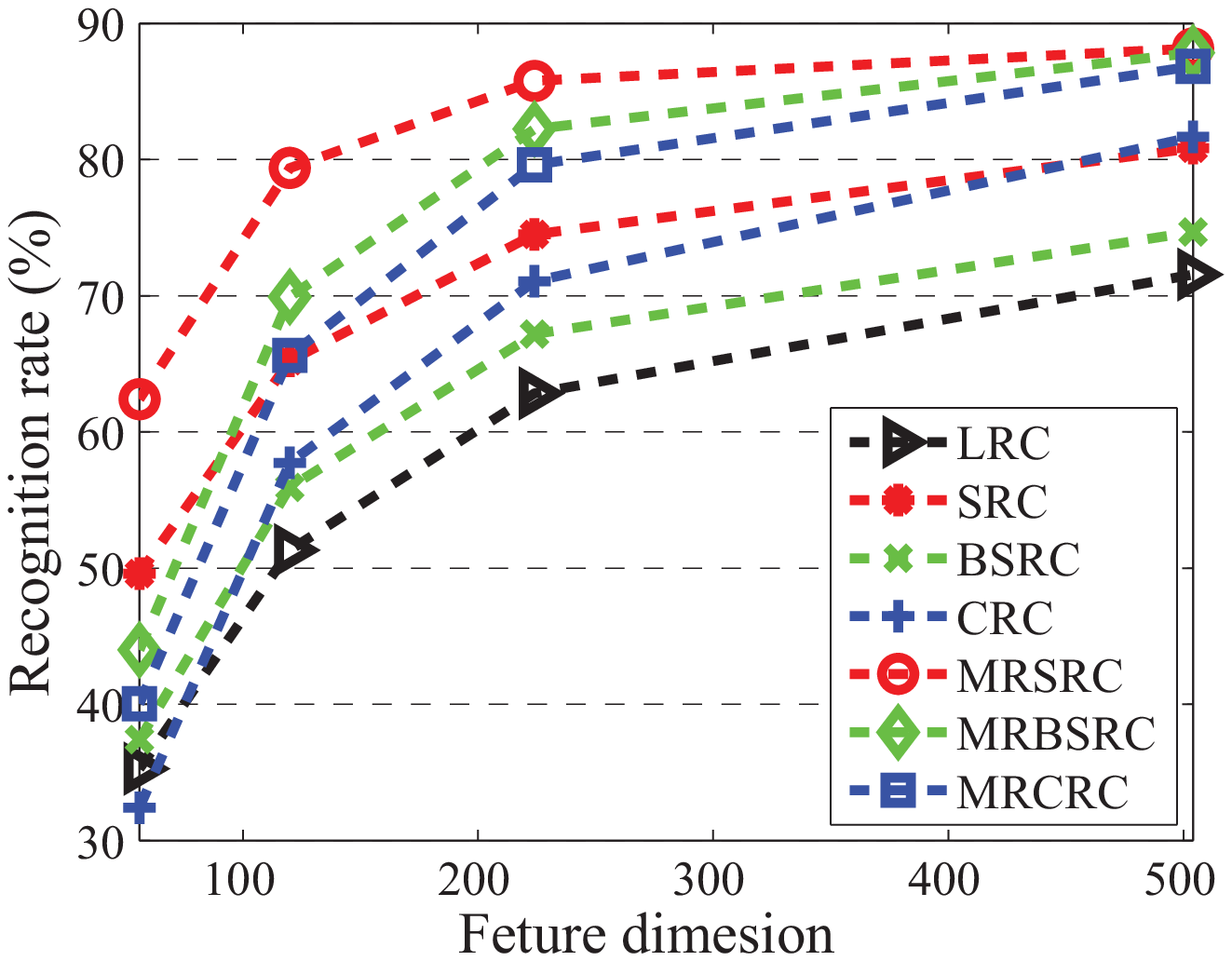}}
 \caption{\textbf{Recognition rates versus feature dimension with random occlusion}.
 Recognition rates as a function of the feature dimension on the Extended Yale B
  Database against 20 percent occlusion with images of the first $K$ subjects.
 (a) $K=10$; (b) $K=20$; (c) $K=38$.}
\label{fig5}
\end{figure*}

\begin{table*}
\caption{Recognition rates (\%) using varying number of training samples per class $n_c$
on the Extended Yale B database with 20 percent occlusion in each test image using ten-run test.
The image size is $24\times21$.
Best results are marked bold.}
\begin{center}
\begin{tabular}
{|c|c|c|c|c|c|c|c|c|}
\hline\cline{1-9}
                      $n_c$      &Methods   &LRC &SRC &BSRC &CRC &MRSRC &MRBSRC &MRCRC
\\ \hline
\multirow{4}{*}{16}              &Mean      &69.30  &75.42  &70.76  &77.29  &\textbf{81.73}  &76.18  &81.17
        \\

                                 &Min~      &67.52  &74.26  &69.57  &75.58  &\textbf{80.26}  &74.26  &79.69
        \\

                                 &Max~      &70.72  &76.32  &72.12  &79.44  &\textbf{82.98}  &77.63  &82.89
        \\
\cline{1-2}
\multirow{4}{*}{20}              &Mean      &70.02  &77.60  &72.25  &79.26  &\textbf{85.36}  &79.38  &82.55
        \\

                                 &Min~      &68.26  &76.81  &71.13  &78.13  &\textbf{83.88}  &77.14  &81.50
        \\

                                 &Max~      &71.46  &78.62  &73.19  &80.43  &\textbf{86.10}  &80.43  &83.47
        \\
\cline{1-2}
\multirow{4}{*}{24}              &Mean      &70.79  &78.93  &73.32  &80.22  &\textbf{87.72}  &82.12  &83.92
        \\

                                 &Min~      &69.82  &77.80  &72.29  &79.36  &\textbf{86.43}  &81.33  &82.65
        \\

                                 &Max~      &72.20  &79.44  &73.93  &80.92  &\textbf{88.98}  &83.47  &84.70
        \\
\cline{1-2}
\multirow{4}{*}{28}              &Mean      &71.04  &79.95  &74.19  &81.04  &\textbf{89.40}  &83.91  &84.36
        \\

                                 &Min~      &70.31  &78.70  &73.03  &80.43  &\textbf{88.65}  &83.14  &83.39
        \\

                                 &Max~      &71.96  &80.59  &74.92  &81.50  &\textbf{89.97}  &85.20  &85.03
        \\ \hline\cline{1-9}
\end{tabular}
\end{center}
\label{tab3}
\end{table*}

\begin{table*}
\caption{Recognition rates (\%) with real-world occlusions on the AR database. 
Some results of LRC, SRC, BSRC and CRC are copied from the corresponding papers.
The image size is $41\times30$.
Best results are marked bold.}
  \centering
\begin{tabular}
{|c|c|c|c|c|c|c|c|}
\hline\cline{1-8}
       &LRC  &SRC &BSRC &CRC &MRSRC &MRBSRC &MRCRC\\
\hline
Sunglasses    &96     &87 &80.5  &68.5  &\textbf{98}  &88.5  &94 \\
Scarves       &95.5    &93.5  &96  &95  &97.5  &95.5  &\textbf{98.5} \\
\hline\cline{1-8}
\end{tabular}
\label{tab4}
\end{table*}

\textbf{Experiment 2--Effect of feature dimension:}
In the second experiment, we study the effect of the feature dimension (or image size)
to the recognition performance using the Extended Yale B database.
To this end, we resize the images to $8\times 7$, $12\times 10$, $16\times 14$,
and $24\times 21$, respectively.
The corresponding downsampling ratio is $1/24$, $1/16$, $1/12$, and $1/8$, respectively.
Fig. \ref{fig5} shows the recognition rates
versus feature dimension with 20 percent occlusion using
facial images of the first $K=10$, $20$ and $38$ subjects.
The three methods in MRARC can improve the corresponding ones in ARC
with varying feature dimensions and number of classes.
In particular, the MRSRC method significantly outperform other
competing methods especially when the feature dimension is low.

\textbf{Experiment 3--Effect of training set size:}
In the third experiment, we analyze the impact of the training set size on
the final face recognition performance with random occlusion.
For each subject, we randomly select $n_c~(=16,20,24,28)$ images for
training and perform recognition on the test images with 20\% occlusion.
We repeat the experiment
ten times and compute the mean, minimum and maximum recognition rates of each algorithm.
The results are reported in Table \ref{tab3}.
It can be found in Table \ref{tab3} that even with small training set,
 MRARC methods can enhance the ARC ones with large margin.

\textbf{Experiment 4--Recognition with real-world occlusions:}
In the fourth experiment, we evaluate the performance of the proposed methods
against real-world occlusions such as sunglass and scarf.
Fig. \ref{fig3}(b) shows four facial images occluded
by sunglasses or scarf in the AR database.
The AR database is used for this experiment
and the images are resized to $41\times 30$.
For each subject, the eight images with varying expressions are utilized for training.
The four images with sunglasses or scarves are used for testing.
Table \ref{tab4} reports the recognition results.
Some results of LRC, SRC, BSRC and CRC are copied
from the corresponding papers.
The results suggest that the proposed
MRARC methods can well handle facial images
with real-world occlusions with
high recognition accuracy.

\begin{figure*}
  \centering
   \subfigure[]{
    \nonumber
   \includegraphics[width=8.0cm]{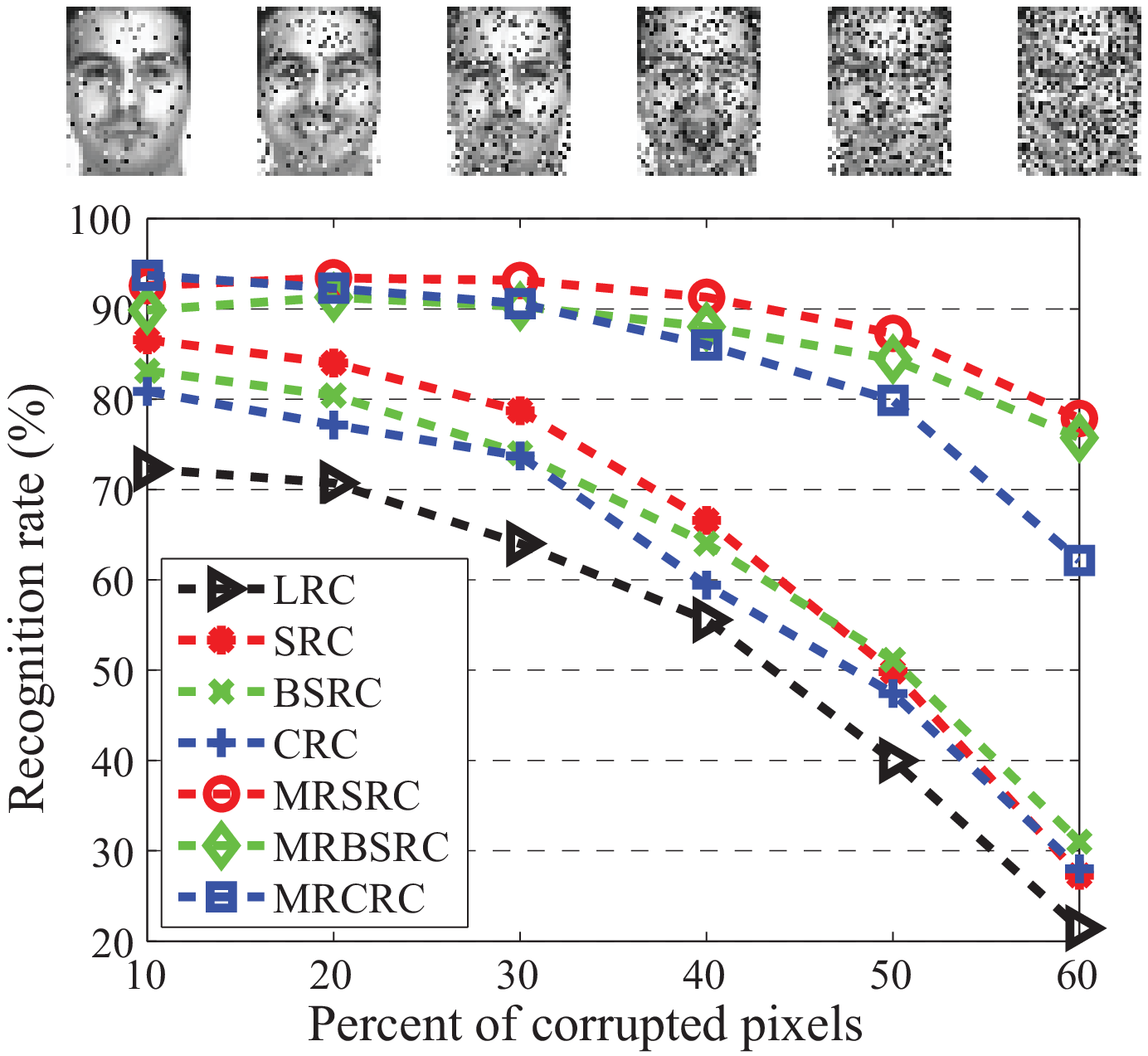}}
   \subfigure[]{
    \nonumber
   \includegraphics[width=8.0cm]{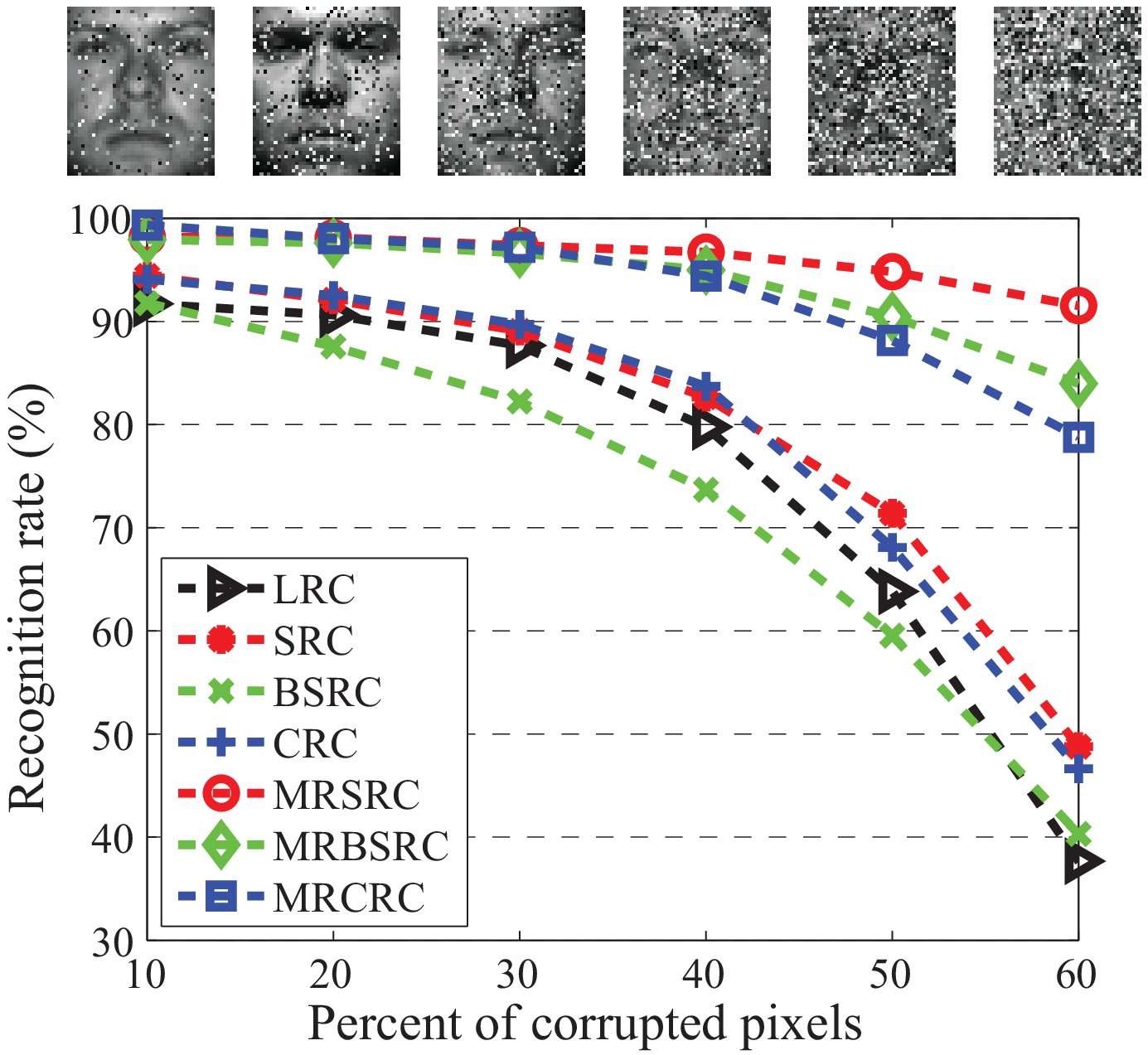}}
 \caption{\textbf{Recognition rates versus percent of corrupted pixels}.
 Recognition rates as a function of the percent of corrupted pixels in each test image.
 (a) the AR database; 
 (b) the Extended Yale B Database.}  
\label{fig6}
\end{figure*}

\subsection{Face Recognition With Corruption}
\label{subsec:RecogCorr}

In this part, we evaluate
the performance of proposed methods for face recognition
and reconstruction against random corruption.

\textbf{Experiment 1--Effect of percent of corruption:}
In the first experiment, we study the performance of proposed methods
against different levels of random pixel corruption.
To this end, a fraction of pixels of each test image are
randomly chosen and their values are replaced
by random values following uniform distribution over the interval $[0, 255]$.
Like the settings in Section \ref{subsec:RecogOcc},
we randomly select half of images per subject in the Extended Yale B databases
for training and the rest are used for testing.
As for the AR database, we use the seven images with expression and illumination variations per subject in the first session for training
and the seven images in the second session for testing.
Fig. \ref{fig6} shows the recognition rates of competing algorithms
as the percent of corruption varies from 10 to 60.
The results demonstrate the superiority of the methods in MRARC over those in ARC
for recognition with random corruption.

\begin{table*}
\caption{Recognition rates (\%) using varying number of training samples per class $n_c$
on the Extended Yale B database with 30 percent corruption in each test image using ten-run test.
The image size is $24\times21$.
Best results are marked bold.}
\begin{center}
\begin{tabular}
{|c|c|c|c|c|c|c|c|c|}
\hline\cline{1-9}

                   $n_c$      &Methods   &LRC &SRC &BSRC &CRC &MRSRC &MRBSRC &MRCRC
\\ \hline
\multirow{4}{*}{16}              &Mean      &71.67  &65.94  &58.50  &59.96  &81.88  &76.77  &\textbf{81.95}
        \\

                                 &Min~      &69.90  &64.80  &56.50  &58.63  &80.35  &75.08  &\textbf{81.25}
        \\

                                 &Max~      &73.03  &66.94  &59.62  &61.43  &\textbf{82.98}  &77.96  &82.89
        \\
\cline{1-2}
\multirow{4}{*}{20}              &Mean      &73.66  &67.17  &56.55  &62.55  &\textbf{85.01}  &79.67  &82.53
        \\

                                 &Min~      &72.78  &65.87  &54.77  &60.36  &\textbf{83.72}  &77.88  &81.17
        \\

                                 &Max~      &74.59  &67.93  &58.39  &63.98  &\textbf{85.94}  &81.25  &83.47
        \\
\cline{1-2}
\multirow{4}{*}{24}              &Mean      &74.38  &67.60  &55.10  &63.03  &\textbf{87.68}  &82.84  &82.86
        \\

                                 &Min~      &73.68  &66.28  &53.54  &62.17  &\textbf{86.60}  &82.07  &82.07
        \\

                                 &Max~      &75.08  &68.59  &56.66  &64.47  &\textbf{88.73}  &85.03  &83.80
        \\
\cline{1-2}
\multirow{4}{*}{28}              &Mean      &74.88  &68.22  &53.17  &63.62  &\textbf{89.35}  &85.21  &83.08
        \\

                                 &Min~      &73.77  &67.68  &52.47  &62.99  &\textbf{88.73}  &84.05  &82.24
        \\

                                 &Max~      &75.66  &69.16  &53.95  &64.80  &\textbf{90.05}  &85.94  &84.05
        \\ \hline\cline{1-9}
\end{tabular}
\end{center}
\label{tab5}
\end{table*}

\textbf{Experiment 2--Effect of training set size:}
In the second experiment, we analyze the impact of the training set size on
the recognition performance with random pixel corruption.
Like the settings in Section \ref{subsec:RecogOcc}, we
vary the number of training samples per class $n_c$ from 16 to 28.
Table \ref{tab5} reports the recognition results
using test images with 30\% random corruption over ten runs.
The results further validate the fact that
MRARC methods can improve the recognition performance of ARC methods
against random corruption in most cases.

\begin{table*}[!t]
\caption{Recognition rates (\%) of various methods on the CMU MoBo database with distinct number $n_f$
of frames per subject  over 10 random runs. Best results are marked bold.}
\begin{center}
\begin{tabular}
{|c|c|c|c|c|c|c|c|c|}
\hline\cline{1-9}

$n_f$              &Methods   &LRC &SRC &BSRC &CRC &MRSRC &MRBSRC &MRCRC
\\ \hline\hline
\multirow{4}{*}{20}&Mean    &56.76  &67.30  &67.57  &69.05  &87.16
                                 &\textbf{87.84}  &87.57
        \\
\cline{2-2}
                   &Min     &52.70  &60.81  &60.81  &64.86  &82.43  &81.08  &\textbf{83.78}
        \\
\cline{2-2}
                   &Max     &62.16  &72.97  &78.38  &78.38  &\textbf{90.54}  &\textbf{90.54}  &\textbf{90.54}
        \\
\cline{1-2}
\multirow{4}{*}{40}&Mean    &61.76  &76.76  &72.30  &70.41  &90.14  &90.27
                                 &\textbf{91.22}
        \\
\cline{2-2}
                   &Min     &52.70  &70.27  &64.86  &59.46  &86.49  &86.49  &\textbf{89.19}
        \\
\cline{2-2}
                   &Max     &75.68  &83.78  &85.14  &81.08  &\textbf{94.59}  &93.24  &\textbf{94.59}
        \\
\cline{1-2}
\multirow{4}{*}{80}&Mean &65.81  &77.30  &73.11  &73.65  &\textbf{92.16}  &91.49  &92.03
        \\
\cline{2-2}
                   &Min     &55.41  &70.27  &63.51  &68.92  &\textbf{87.84}  &86.49  &\textbf{87.84}
        \\
\cline{2-2}
                   &Max    &77.03  &82.43  &81.08  &78.38  &\textbf{95.95}  &94.59  &94.59
        \\ \hline
\end{tabular}
\end{center}
\label{tab6}
\end{table*}

\subsection{Results on the CMU MoBo Database}
\label{subsec:MoBo}
In this part, we evaluate the performance of  MRARC for
image set based face recognition (ISFR)
using the CMU Mobo database \cite{Gross01}.
The database consists of 96 video sequences of 24 subjects walking on a treadmill.
For each subject, 4 video sequences are taken with four distinct walking patterns, respectively.
The detected facial images using the Viola-Jones face detector \cite{Viola04} are resized to $40\times 40$.
For each facial image, we extract the Local Binary Pattern (LBP)
features \cite{Ahonen06} and normalize each feature to have unit Euclidean norm.
To evaluate the robustness of proposed methods against corruption,
we randomly choose 10 percent of the entries of each feature vector and replace them
by random values following the uniform distribution $[0, 1]$.
In the experiment, we randomly select a video sequence for training and use the rest for testing.
To extend ARC and MRARC methods for ISFR, we
use the average class-dependent reconstruction residuals of the images
in the query set for recognition \cite{Zhu14}.
Eventually, the query set is assigned to the class minimizing the average residual.
Instead of using all frames of each video,
we randomly choose $n_f~(=20, 40, 80)$ frames
of images from each video and use them to construct the training and query (testing) set for recognition.
To obtain reliable results, we repeat each test ten times and compute the average, minimal and
maximum recognition rates among the ten tests.
Table \ref{tab6} reports the recognition results.
The results show that the three MRARC methods have close performance
and outperform other competing RC methods.

\subsection{Multimodal Face Recognition}
\label{subsec:MultimodalR}
In this section, we compare the MRJSRC method in the MRARC framework
with JSRC \cite{Shekhar14} for multimodal face recognition.
For the comparison between JSRC and other multimodal recognition methods,
see the  reference \cite{Shekhar14}.

\begin{figure}[!t]
  \centering
  \subfigure[]{
    \nonumber
   \includegraphics[height=2.0cm]{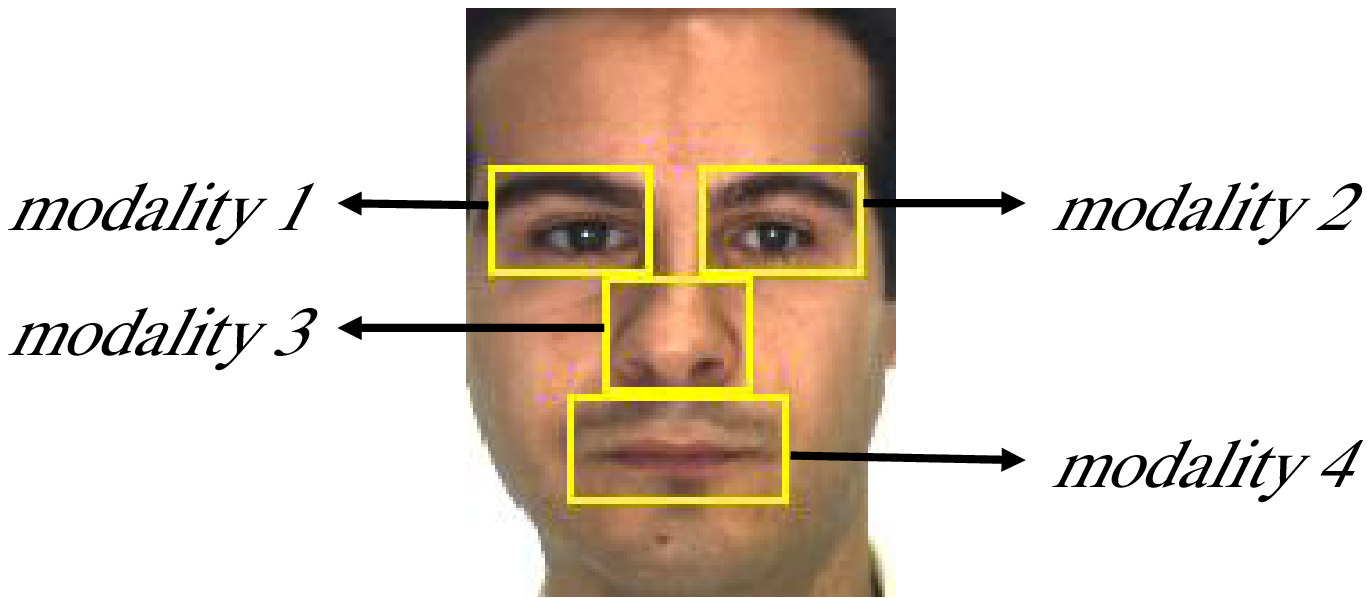}}
   \hspace{0.3in}
   \subfigure[]{
    \nonumber
   \includegraphics[height=2.0cm]{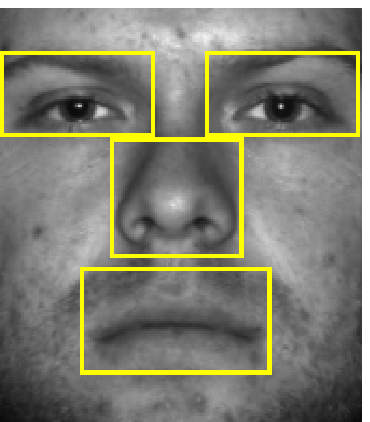}}
 \caption{\textbf{Extracted four modalities from an image. }
 (a) AR database and (b) Extended Yale B (EYaleB) database.}
\label{fig7}
\end{figure}

In the first experiment, we extract four weak modalities from each facial image
for evaluation \cite{Shekhar14}.
They are the left and right periocular, nose, and mouth as shown in Fig. \ref{fig7}.
For the AR database, the images are resized to $41\times 30$.
Like the settings in Experiment 1 in Section \ref{subsec:RecogCorr},
we use the seven images with expression and illumination variations per subject in the first
session for training and the seven images in the second session for testing.
For the EYaleB database, the images are resized to $24\times 21$.
We randomly select half of all images per subject for training and the rest for testing.
Analogously, we use the original intensity values of the images for the experiments.

\begin{table}
\caption{Recognition rates (\%) using distinct
number of modalities. Modalities include
1. left periocular, 2. right periocular, 3. nose, and 4. mouth.
EYaleB+10\%Occlu means that each test image in the EYaleB database
has 10\% region occluded by a baboon image as in Fig. \ref{fig4}.
Best results are marked bold. }
\begin{center}
\begin{tabular}
{|c|c|c|c|c|c|}
\hline\cline{1-6}
Database &Algorithm         &\{1\} &\{1,2\} &\{1,2,3\} &\{1,2,3,4\}\\
\hline
\multirow{2}{*}{AR} &JSRC        &73.14 &82.14  &86.43  &89.71  \\
&MRJSRC      &\textbf{78.71}  &\textbf{84.71}  &\textbf{89.43}  &\textbf{91.14}  \\
\cline{1-6}

\multirow{2}{*}{EYaleB} &JSRC        &83.06  &95.89  &96.88  &97.86  \\
&MRJSRC      &\textbf{85.36}  &\textbf{97.70}  &\textbf{97.78}  &\textbf{98.68}  \\
\cline{1-6}

\multirow{2}{*}{EYaleB+10\%Occlu} &JSRC        &69.98  &88.90  &90.38  &95.23  \\
&MRJSRC      &\textbf{71.13}  &\textbf{89.56}  &\textbf{92.35}  &\textbf{95.89}  \\
\cline{1-6}

\multirow{2}{*}{EYaleB+20\%Occlu} &JSRC        &63.08  &82.24  &83.55  &91.37  \\
&MRJSRC      &\textbf{64.80}  &\textbf{84.62}  &\textbf{85.86}  &\textbf{92.43}  \\
\hline\cline{1-6}
\end{tabular}
\end{center}
\label{tab7}
\end{table}

Table \ref{tab8} reports the recognition results
of MRJSRC and JSRC using varying number of modalities.
Based on the results, we can draw the following conclusions.
\begin{itemize}
\item First, the recognition rate of each competing method grows rapidly
as the number of modalities increases from 1 to 4.
This suggests that both JSRC and MRJSRC can exploit the complementarity among
distinct modalities to improve the recognition performance.
However, the proposed MRJSRC method stably improves JSRC with varying number of
modalities.

\item Second,  the competing methods using more modalities have better robustness
against random block occlusion on the EYaleB database.
For example, the recognition rate of MRJSRC using the single modality 1
(i.e., the left periocular modality)
 varies from $85.36\%$ without occlusion to $64.80\%$ with $20\%$ occlusion, declining over $20\%$.
In contrast, the recognition rate of MRJSRC using the four modalities varies from
$98.68\%$ to $92.43\%$, declining less than $7\%$.

\end{itemize}

\begin{figure}[!t]
  \centering
   \subfigure[View 1]{
    \nonumber
   \includegraphics[height=1.4cm]{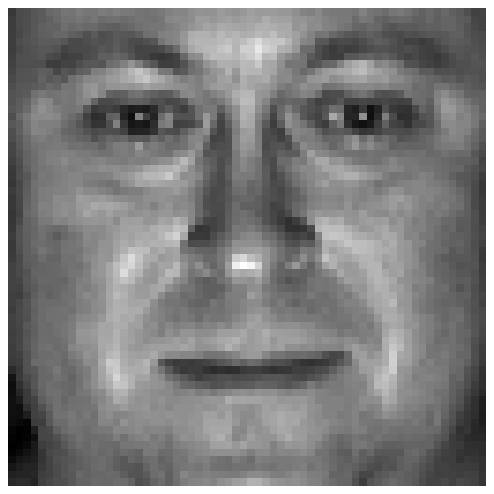}
   \includegraphics[height=1.4cm]{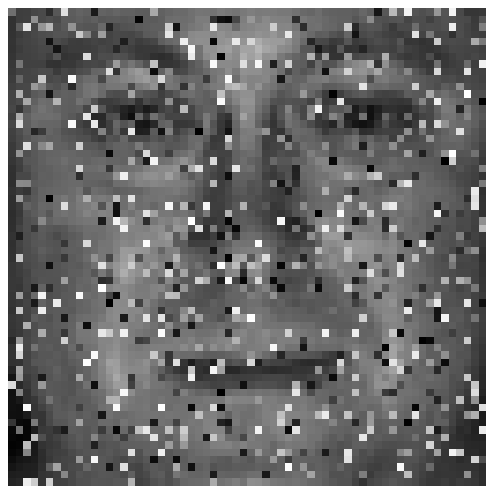}}
   \hspace{0.2in}
   \subfigure[View 2]{
    \nonumber
   \includegraphics[height=1.4cm]{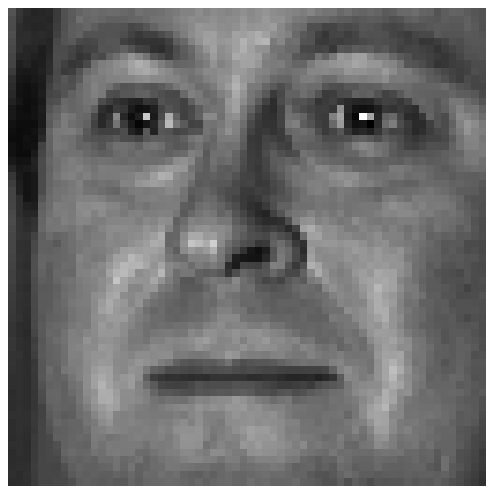}
   \includegraphics[height=1.4cm]{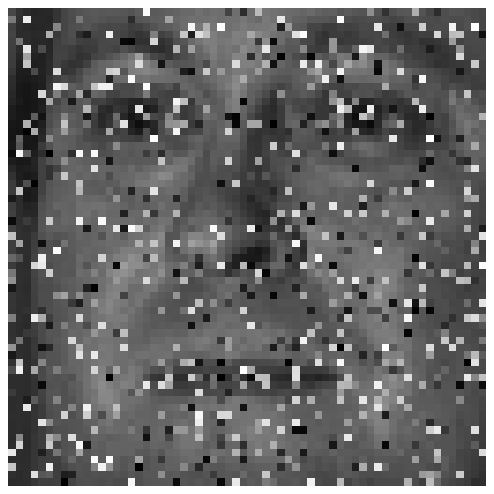}}
 \caption{\textbf{Facial images in two views of a subject in the CMU PIE database.}
  (a) View 1; (b) View 2. In each subfigure,
  from left to right: the original image and
  the noisy image with 20\% random corruption. }
\label{fig8}
\end{figure}

\begin{figure}[!t]
  \centering
   \includegraphics[width=8.0cm]{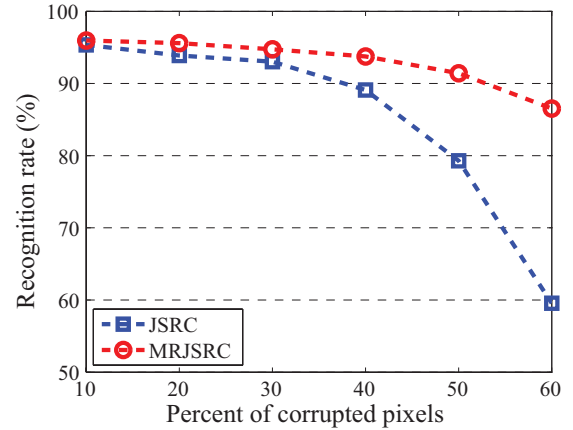}
 \caption{\textbf{Recognition rates as a function of the percent of corrupted pixels using
 two views in the CMU PIE database.}}
\label{fig9}
\end{figure}

In the second experiment, we evaluate the performance of the proposed method
for multiview face recognition using the CMU PIE database \cite{Sim03}.
This database consists of 41,368 images
of 68 people, of which the facial images are captured under 13 different poses, 43
different illumination conditions, and with 4 different expressions.
Two near frontal poses (i.e., C09 and C29) are selected to construct the
multiview setting, as shown in Fig. \ref{fig8}.
Thus, a pair of images of one subject under the two poses are  regarded
as a two-view or two-modality sample.
Each image is resized to $32\times32$.
In the experiment, we randomly select half of all images per subject for
training and the rest for testing.
As shown in Fig. \ref{fig8},  a fraction of pixels of each test image are
randomly selected and their values are replaced
by random values following uniform distribution over the interval $[0, 255]$.

Fig. \ref{fig9} shows the recognition rates of competing methods as a function
of the percent of corrupted pixels in each test image.
We also compare JSRC and MRJSRC using single-view and two-view samples and
with varying percent of random corruption. The results are reported in Table \ref{tab8}.
Concretely, View 1 and View 2 correspond to Pose C09 and Pose C29, respectively.
From Fig. \ref{fig9} and Table \ref{tab8},
we can find that the proposed MRJSRC method stably outperforms JSRC
in varying number of views and corruption level.
This comes from the fact that MRJSRC has decent robust property
and can well take advantage of the complementary information
among multiple modalities.

\begin{table}
\caption{Recognition rates (\%) using distinct number of
views with varying percent $p_c$ of randomly corrupted pixels.
Best results are marked bold. }
\begin{center}
\begin{tabular}
{|c|c|c|c|c|}
\hline\cline{1-5}
$p_c(\%)$  &Algorithm         &View 1 $+$ View 2 &View 1 &View 2 \\
\hline
\multirow{2}{*}{10} &JSRC        &95.34  &93.26  &92.52  \\
&MRJSRC                          &\textbf{95.96}  &\textbf{94.00}  &\textbf{93.01}  \\
\cline{1-5}

\multirow{2}{*}{20} &JSRC        &93.87  &91.91  &90.69  \\
&MRJSRC                          &\textbf{95.59}  &\textbf{93.75}  &\textbf{92.28}  \\
\cline{1-5}

\multirow{2}{*}{30} &JSRC        &93.01  &89.22  &86.76  \\
&MRJSRC                          &\textbf{94.73}  &\textbf{93.63}  &\textbf{91.30}  \\
\cline{1-5}

\multirow{2}{*}{40} &JSRC        &89.09  &82.23  &78.55  \\
&MRJSRC                          &\textbf{93.75}  &\textbf{91.42}  &\textbf{89.95}  \\
\cline{1-5}

\multirow{2}{*}{50} &JSRC        &79.29  &68.75  &59.93  \\
&MRJSRC                          &\textbf{91.42}  &\textbf{87.87}  &\textbf{84.56}  \\
\cline{1-5}

\multirow{2}{*}{60} &JSRC        &59.56  &45.71  &34.93  \\
&MRJSRC                          &\textbf{86.52}  &\textbf{76.35}  &\textbf{71.69}  \\
\hline\cline{1-5}
\end{tabular}
\end{center}
\label{tab8}
\end{table}


\section{Conclusion}
\label{sec:Conclusion}

This paper presents a novel general classification framework termed as MRARC
for robust face recognition and reconstruction.
The proposed MRARC framework is based on the modal regression and
does not require the noise to follow any specific distribution.
This gives rise to the ability of MRARC in handling
various complicated noises in reality.
Using MRARC as a platform, we have also developed several novel RC methods
for robust unimodal and multimodal face recognition.
The experiments on real-world databases show the efficacy of the proposed methods
for robust face recognition and reconstruction.



\end{document}